\pdfoutput=1
\RequirePackage{etex}
\documentclass[11pt]{article}
\usepackage[preprint]{acl}
\usepackage{times}
\usepackage{latexsym}

\usepackage[T1]{fontenc}
\usepackage[utf8]{inputenc}

\usepackage{natbib}
\usepackage{tcolorbox}
\usepackage{tcolorbox}
\tcbuselibrary{breakable}
\usepackage{linegoal}
\usepackage{color}
\usepackage{tabularx}
\usepackage{tcolorbox}
\usepackage{xcolor}
\usepackage{colortbl}
\usepackage{tabu}
\usepackage{booktabs}
\usepackage{multirow}

\usepackage{graphicx} 
\usepackage{float}
\usepackage{subfigure} 
\usepackage{amssymb}
\usepackage{tablefootnote}
\usepackage{xspace}

\usepackage{float}
\usepackage{amsmath}
\usepackage{bm}
\usepackage{algorithmicx}
\usepackage{algorithm}
\usepackage{algpseudocode}

\usepackage{arydshln}

\usepackage{amssymb}
\usepackage{pifont}
\usepackage{enumitem}
\usepackage{fontawesome5}
\usepackage{academicons}
\definecolor{C_SECOND_SOFT}{rgb}{0.9, 0.9, 0.9}  % 示例颜色
\definecolor{C_SECOND}{rgb}{0.0, 0.4, 0.8}      % 示例颜色

\usepackage{listings}
\usepackage{microtype}

\usepackage{inconsolata}

\usepackage{graphicx}

\usepackage[mathscr]{euscript}
\usepackage{silence}
\usepackage{enumitem}
\usepackage[utf8]{inputenc}
\usepackage[T1]{fontenc}
\usepackage{hyperref}
\usepackage{url}
\usepackage{booktabs}
\usepackage{amsfonts}
\usepackage{nicefrac}
\usepackage{microtype}
\usepackage{xcolor}
\usepackage{wrapfig}
\usepackage{lipsum}
\usepackage{inconsolata}
\usepackage{colortbl}
\usepackage{makecell}
\usepackage[linewidth=1pt]{mdframed}
\usepackage{algorithm}
\usepackage[algo2e,ruled,linesnumbered,noline,noend]{algorithm2e}
\usepackage{float}
\usepackage{times}
\usepackage{latexsym}
\usepackage{multirow}
\usepackage{pifont}
\usepackage{soul}
\usepackage{comment}
\usepackage[export]{adjustbox}
\usepackage{csquotes}
\usepackage{standalone}
\usepackage{amsmath}
\usepackage{amssymb}
\usepackage{amsthm}
\usepackage{graphicx}
\usepackage{subcaption}
\usepackage{array}
\usepackage{tabu}
\usepackage{cases}
\usepackage{diagbox}
\usepackage{verbatim}
\usepackage{tabulary}
\usepackage[mathscr]{euscript}
\usepackage{mathtools}
\usepackage{algpseudocode}
\usepackage{stmaryrd}
\usepackage{tikz-dependency}
\usetikzlibrary{automata,decorations.markings,arrows,positioning,matrix,calc,patterns,angles,quotes,calc}
\usepackage{adjustbox}
\usepackage{tabularx}
\usepackage{xspace}
\usepackage{afterpage}
\usepackage{bm}
\usepackage{longtable}
\usepackage{color}
\usepackage{slashbox}
\usepackage[toc,page]{appendix}
\usepackage{boldline}
\usepackage[shortcuts]{extdash}
\usepackage{blindtext}
\usepackage{caption}
\usepackage{capt-of}
\usepackage{varwidth}
\usepackage{listings}
\usepackage{pythonhighlight}
\usepackage{graphics}
\usepackage{newfloat}
\usepackage{epigraph}
\allowdisplaybreaks[4]
\definecolor{orange}{rgb}{1,0.5,0}
\definecolor{mdgreen}{rgb}{0.05,0.6,0.05}
\definecolor{mdblue}{rgb}{0,0,0.7}
\definecolor{dkblue}{rgb}{0,0,0.5}
\definecolor{dkgray}{rgb}{0.3,0.3,0.3}
\definecolor{slate}{rgb}{0.25,0.25,0.4}
\definecolor{gray}{rgb}{0.5,0.5,0.5}
\definecolor{ltgray}{rgb}{0.7,0.7,0.7}
\definecolor{purple}{rgb}{0.7,0,1.0}
\definecolor{lavender}{rgb}{0.65,0.55,1.0}
\definecolor{mypurple}{RGB}{111,61,121}
\definecolor{myblue}{RGB}{46,88,180}
\definecolor{myred}{RGB}{181,68,106}
\definecolor{myyellow}{RGB}{204,143,55}
\definecolor{deepgreen}{rgb}{0.0, 0.5, 0.0}

\newcommand{\hlred}{\colorlet{c}{red!20}\sethlcolor{c}\hl}
\newcommand{\hlgreen}{\colorlet{c}{green!20}\sethlcolor{c}\hl}

\newcommand{\ours}{FinanceReasoning\xspace}

\newcommand{\narticle}{6,138\xspace}
\newcommand{\nfunction}{3,133\xspace}

\newcommand{\eg}{\hbox{\emph{e.g.,}}\xspace}
\newcommand{\ie}{\hbox{\emph{i.e.,}}\xspace}

\newcommand{\up}[1]{\textcolor{red}{(+#1)}}
\newcommand{\down}[1]{\textcolor{blue}{(-#1)}}

\newcommand{\homepage}{\raisebox{-1.5pt}{\includegraphics[height=1.2em]{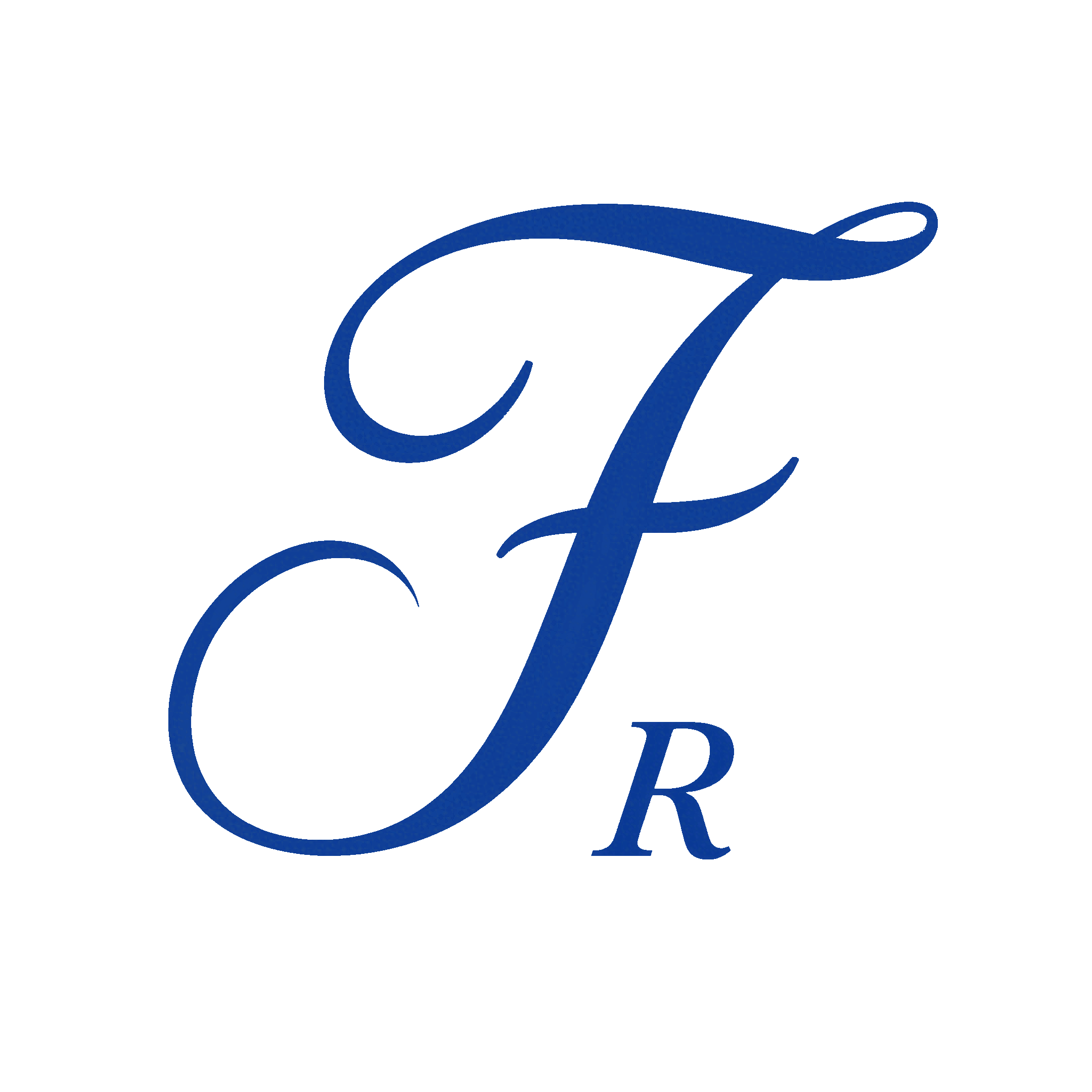}}}
\newcommand{\github}{\raisebox{-1.5pt}{\includegraphics[height=1em]{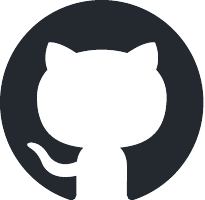}}}
\newcommand{\huggingface}{\raisebox{-1.5pt}{\includegraphics[height=1em]{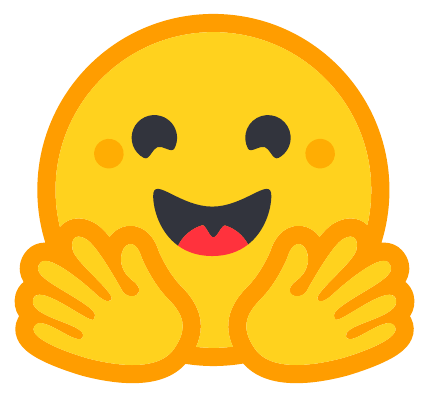}}}

\newcommand{\logo}{%
    \raisebox{-0.7ex}{% 微调垂直位置
        \includegraphics[width=2em,height=2em,keepaspectratio]{figures/blue_logo.png}%
    }%
}

\title{\logo\ours: Benchmarking Financial Numerical Reasoning \\More Credible, Comprehensive and Challenging}

\author{Author 1 \and ... \and Author n \\
        Address line \\ ... \\ Address line}
\author{
Zichen Tang \quad 
Haihong E\thanks{Corresponding author.} \quad 
\bf{Ziyan Ma} \quad 
\bf{Haoyang He} \quad
\bf{Jiacheng Liu} \\ 
\bf{Zhongjun Yang} \quad
\bf{Zihua Rong} \quad 
\bf{Rongjin Li} \quad
\bf{Kun Ji} \quad 
\bf{Qing Huang} \quad
\bf{Xinyang Hu} \\
\bf{Yang Liu} \quad 
\bf{Qianhe Zheng} \vspace{4pt}\\
Beijing University of Posts and Telecommunications\\\newline \vspace{10pt}
}

\begin{document}
\maketitle

% \begin{minipage}[t]{2\linewidth}
% \vspace{-2.0cm}
%   \centering
%   \github \href{https://github.com/yale-nlp/FinanceMath}{\xspace\texttt{github.com/yale-nlp/FinanceMath}} \\\vspace{2pt}
%   \homepage \href{https://financemath-acl2024.github.io}{\xspace\texttt{financemath-acl2024.github.io}} \\\vspace{2pt}
%   \huggingface \href{https://huggingface.co/FinanceMath}{\xspace\texttt{huggingface.co/FinanceMath}} \\
% \vspace{0.5cm}
% \end{minipage}

\begin{minipage}[t]{2\linewidth}
\vspace{-1.6cm}
  \centering
  \homepage~~\href{https://bupt-reasoning-lab.github.io/FinanceReasoning}{\texttt{bupt-reasoning-lab.github.io/FinanceReasoning}} \\
  \github~~\href{https://github.com/BUPT-Reasoning-Lab/FinanceReasoning}{\texttt{BUPT-Reasoning-Lab/FinanceReasoning}} 
  \hfill
  \huggingface~~\href{https://huggingface.co/datasets/BUPT-Reasoning-Lab/FinanceReasoning}{\texttt{BUPT-Reasoning-Lab/FinanceReasoning}} \\
\vspace{0.5cm}
\end{minipage}

\begin{abstract}
We introduce \textbf{FinanceReasoning}, a novel benchmark designed to evaluate the reasoning capabilities of large reasoning models (LRMs) in financial numerical reasoning problems. Compared to existing benchmarks, our work provides three key advancements. (1) \textbf{Credibility}: We update 15.6\% of the questions from four public datasets, annotating 908 new questions with detailed Python solutions and rigorously refining evaluation standards. This enables an accurate assessment of the reasoning improvements of LRMs. (2) \textbf{Comprehensiveness}: FinanceReasoning covers 67.8\% of financial concepts and formulas, significantly surpassing existing datasets. Additionally, we construct 3,133 Python-formatted functions, which enhances LRMs' financial reasoning capabilities through refined knowledge (\eg{ 83.2\% $\rightarrow$ 91.6\% for GPT-4o}). (3) \textbf{Challenge}: Models are required to apply multiple financial formulas for precise numerical reasoning on 238 \emph{Hard} problems. The best-performing model (\ie{ OpenAI o1 with PoT}) achieves 89.1\% accuracy, yet LRMs still face challenges in numerical precision. We demonstrate that combining Reasoner and Programmer models can effectively enhance LRMs' performance (\eg{ 83.2\% $\rightarrow$ 87.8\% for DeepSeek-R1}). Our work paves the way for future research on evaluating and improving LRMs in domain-specific complex reasoning tasks.

\end{abstract}

\begin{figure}[!t]
    \centering
    \includegraphics[width = \linewidth]{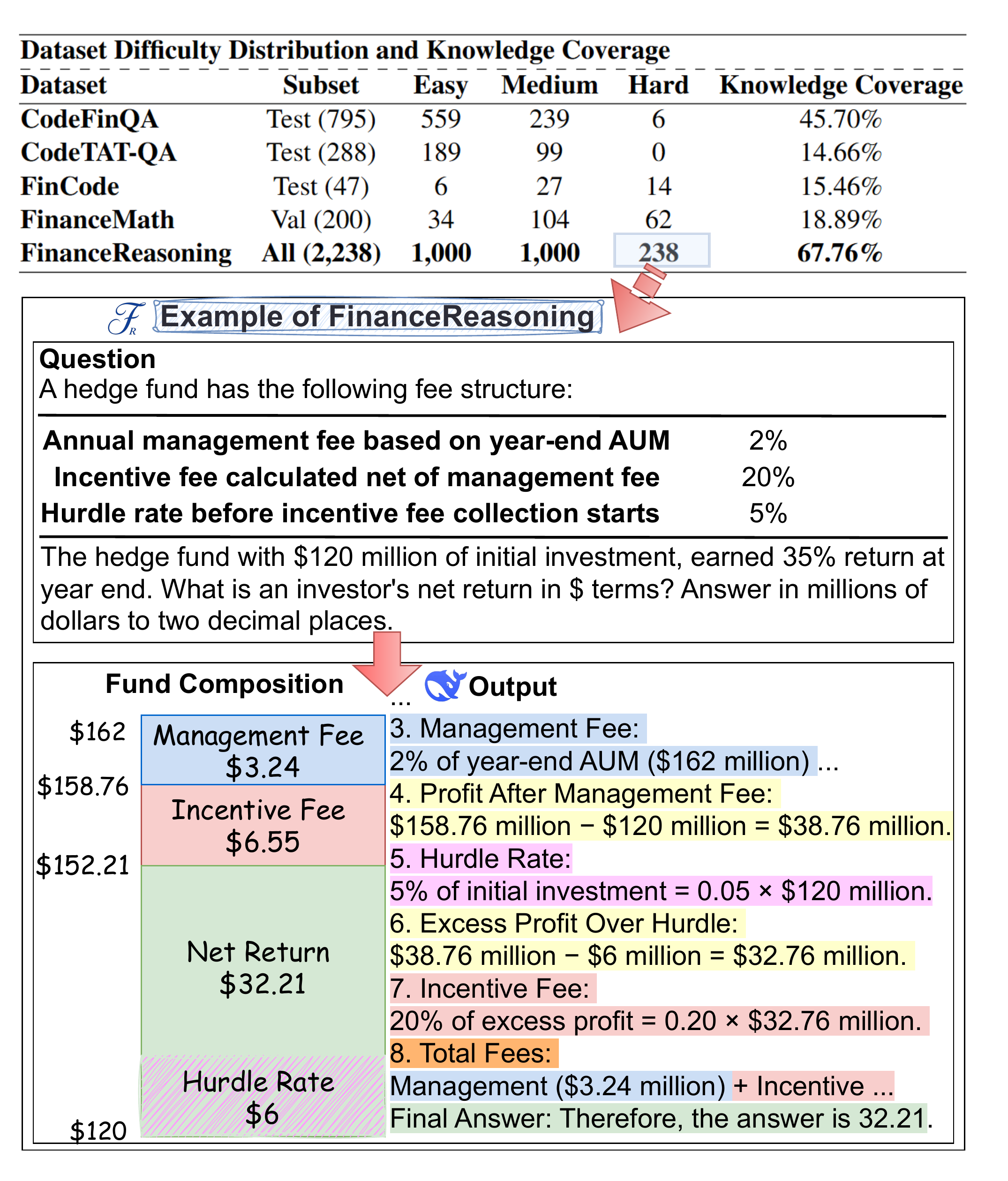}
    \caption{Statistics and an example of FinanceReasoning. The Knowledge Coverage is calculated as the proportion of financial calculations involved in the questions relative to the financial encyclopedia. To address the given problem, LRMs are required to first select appropriate financial formulas based on the given conditions (\eg{ hurdle rate}) and perform step-by-step precise numerical computations with rounding requirements.}
    \label{fig:example}
\end{figure}

% Furthermore, LLMs must identify numerical data from both textual and tabular sources and perform multi-step calculations precisely to determine the investor's net return.

% To address the given problem, LRMs are required to first select appropriate financial formulas based on the given conditions (\eg{ hurdle rate}), as various similar calculation methods exist for different constraints. Subsequently, the model must perform step-by-step precise numerical computations and provide the net return rounded to two decimal places.
\begin{table*}
\centering

\renewcommand{\arraystretch}{1.2}

\resizebox{\textwidth}{!}{
    \tiny
    \begin{tabular}{c c c c c}
    \hline
    
    \multicolumn{5}{l}{\textbf{(a) Re-annotation}} \\ \hdashline
    \textbf{Dataset} & Subset & Answer Corrections & Question Disambiguations  & Total \\ \hline

%\multicolumn{5}{l}{\textbf{(a) Re-annotation}} \\     \hdashline
%    \multirow{2}{*}{\rule{0pt}{1.2ex}\textbf{Dataset}} & 
%    \multirow{2}{*}{\rule{0pt}{1.2ex}Subset} & 
%    \multirow{2}{*}{\rule{0pt}{1.2ex}Answer Corrections} & 
%    \multirow{2}{*}{\rule{0pt}{1.2ex}Question Disambiguations} & 
%    \multirow{2}{*}{\rule{0pt}{1.2ex}Total} \\
%    & & & & \\ 
     \textbf{CodeFinQA} & Test (795) & 55 (6.92\%) & 58 (7.30\%) & 113 (14.21\%) \\ 
     \textbf{CodeTAT-QA} & Test (288) & 19 (6.60\%) & 9 (3.13\%) & 28 (9.72\%) \\ 
     \textbf{FinCode} & Test (47) & 6 (12.77\%) & 1 (2.13\%) & 7 (14.89\%) \\ 
        \textbf{FinanceMath} & Val (200) & 15 (7.50\%) & 45 (22.50\%) & 60 (30.00\%) \\ 
    \hline
    \multicolumn{5}{l}{\textbf{(b) Re-evaluation}} \\ \hdashline
    \textbf{Dataset (Criteria)} & DeepSeek-V3 & DeepSeek-R1 & $\Delta$ (R1 - V3) &  \% Relative Increase (R1 vs. V3) \\ \hline
     \textbf{CodeFinQA (Silver)} & \textbf{61.76} & 60.88 & \cellcolor{red!35}-0.88 & \cellcolor{red!50}-1.42\% \\ 
      \textbf{CodeFinQA (Gold)} & \cellcolor{yellow!25}85.41 & \cellcolor{yellow!45}\textbf{87.42} & \cellcolor{yellow!25}2.01 & \cellcolor{yellow!45}2.35\% \\ 
      \textbf{CodeTAT-QA (Silver)} & 89.24 & \textbf{89.58} & 0.34 & 0.38\% \\ 
        \textbf{CodeTAT-QA (Gold)} & \cellcolor{yellow!45}91.67 & \cellcolor{yellow!55}\textbf{93.75} & \cellcolor{yellow!25}2.08 & \cellcolor{yellow!35}2.27\% \\ 
        \textbf{FinCode (Silver)} & 80.85 & \textbf{82.98} & 2.13 & 2.63\% \\ 
        \textbf{FinCode (Gold)} & \cellcolor{yellow!30}87.72 & \cellcolor{yellow!60}\textbf{95.74} & \cellcolor{yellow!45}8.02 & \cellcolor{yellow!50}9.14\% \\ 
        \textbf{FinanceMath (Silver)} & 58.50 & \textbf{71.00} & 12.50 & 21.37\% \\ 
        \textbf{FinanceMath (Gold)} & \cellcolor{yellow!20}59.50 & \cellcolor{yellow!40}\textbf{83.50} & \cellcolor{yellow!45}24.00 & \cellcolor{yellow!50}40.34\% \\  \hline
    \end{tabular}
}

\caption{(a) \textbf{Re-annotation}: For the \emph{test} or \emph{validation} sets of four datasets, the proportion of updated examples ranges from 9.72\% to 30.00\%. (b) \textbf{Re-evaluation}: \textbf{Silver} denotes the Accuracy on the original dataset under relaxed criteria, while \textbf{Gold} represents the results on the re-annotated dataset under strict criteria. Rigorous revision and evaluation reveal LRMs' actual performance and DeepSeek-R1's significant improvement over DeepSeek-V3.}

\label{tab:Table1}

\end{table*}
\section{Introduction}
\epigraph{``\textit{If you cannot measure it, you cannot improve it.}''}{--- Lord Kelvin}

Recently, combined with train-time scaling and test-time scaling~\cite{kaplan2020scaling,openai2024learning}, large language models (LLMs) have exhibited remarkable reasoning capabilities~\cite{xu2025towards}, through a long reasoning process and effective reasoning strategies. These reasoning-enhanced models (\ie Large reasoning models (LRMs))~\cite{openai2024o1,openai2024o1mini,openai2025o3mini,guo2025deepseekr1,qwq-32b-preview,team2025kimi,geminiteam2025gemini}, are able to tackle complex tasks that require multi-step reasoning, 
such as code~\cite{jain2024livecodebench, chen2021humaneval}, math~\cite{mao-etal-2024-champ, lightman2023let}, and science~\cite{lu2023mathvista, yue2023mmmu, wang2024scibench}.

However, as illustrated in \autoref{fig:example}, more real-world domain-specific numerical reasoning tasks (\eg financial quantitative analysis) challenge LRMs to deeply understand and apply domain-specific knowledge, and perform intricate mathematical calculations based on hybrid contexts such as table and text~\cite{Plaat2024ReasoningWL,chen-etal-2023-theoremqa,wang-zhao-2024-metacognitive,DBLP:journals/nature/RomeraParedesBNBKDREWFKF24}. Specifically, in the high-stakes financial domain, where precision and transparent reasoning are paramount~\cite{krumdick-etal-2024-bizbench}, the reasoning capabilities of LRMs must be further validated and accurately assessed. Existing numerical reasoning benchmarks for finance are limited in their notation quality, coverage of specific knowledge in the financial domain, and complexity of reasoning~\cite{chen-etal-2021-finqa, chen-etal-2022-convfinqa, zhu-etal-2021-tat, zhao-etal-2024-knowledgefmath, krumdick-etal-2024-bizbench}. As illustrated in \autoref{tab:Table1}, DeepSeek-R1 have achieved greater accuracy 90\% in easier datasets and are saturated due to annotation quality, making it difficult to objectively evaluate their actual reasoning capabilities and analyze their shortcomings.

Therefore, we propose \textbf{FinanceReasoning}, a credible, comprehensive, and challenging financial numerical reasoning benchmark to evaluate the reasoning capabilities of LRMs in the finance domain. The dataset comprises a total of 2,238 problems covering diverse financial knowledge, of which 1,420 problems have been reviewed and revised based on public datasets, while 908 problems were automatically generated by LLM (\ie GPT-4o) and subsequently annotated by experts. Each problem includes hybrid contexts, unambiguous questions, Python-formatted solutions, and precise answers, providing a reliable reference for accurately evaluating the complex numerical reasoning capabilities of LRMs. Additionally, we have collected and open-sourced a comprehensive financial function library containing 3,133 Python-formatted functions. Each function includes precise functional descriptions, parameter explanations, and step-by-step implementation code, offering a high-quality structured knowledge base to automatically build domain-specific reasoning problems and enhance LLMs' domain-specific reasoning capabilities through knowledge retrieval.

We evaluate six current open-source and proprietary LRMs~\cite{openai2024o1,openai2024o1mini,openai2025o3mini,guo2025deepseekr1,qwq-32b-preview,geminiteam2025gemini}, using Chain-of-Thought (CoT)~\cite{wei2022chain} and Program-of-Thought (PoT)~\cite{chen2023program}. We also evaluate seven LLMs without reasoning-specific enhancement~
\cite{geminiteam2025gemini,openai2024gpt4o,claude3.5sonnet,deepseekai2024deepseekv3,llama3.3modelcard,llama3.1modelcard,qwen2025qwen25}.

\begin{figure*}[!t]
    \centering
    \includegraphics[width = \linewidth]{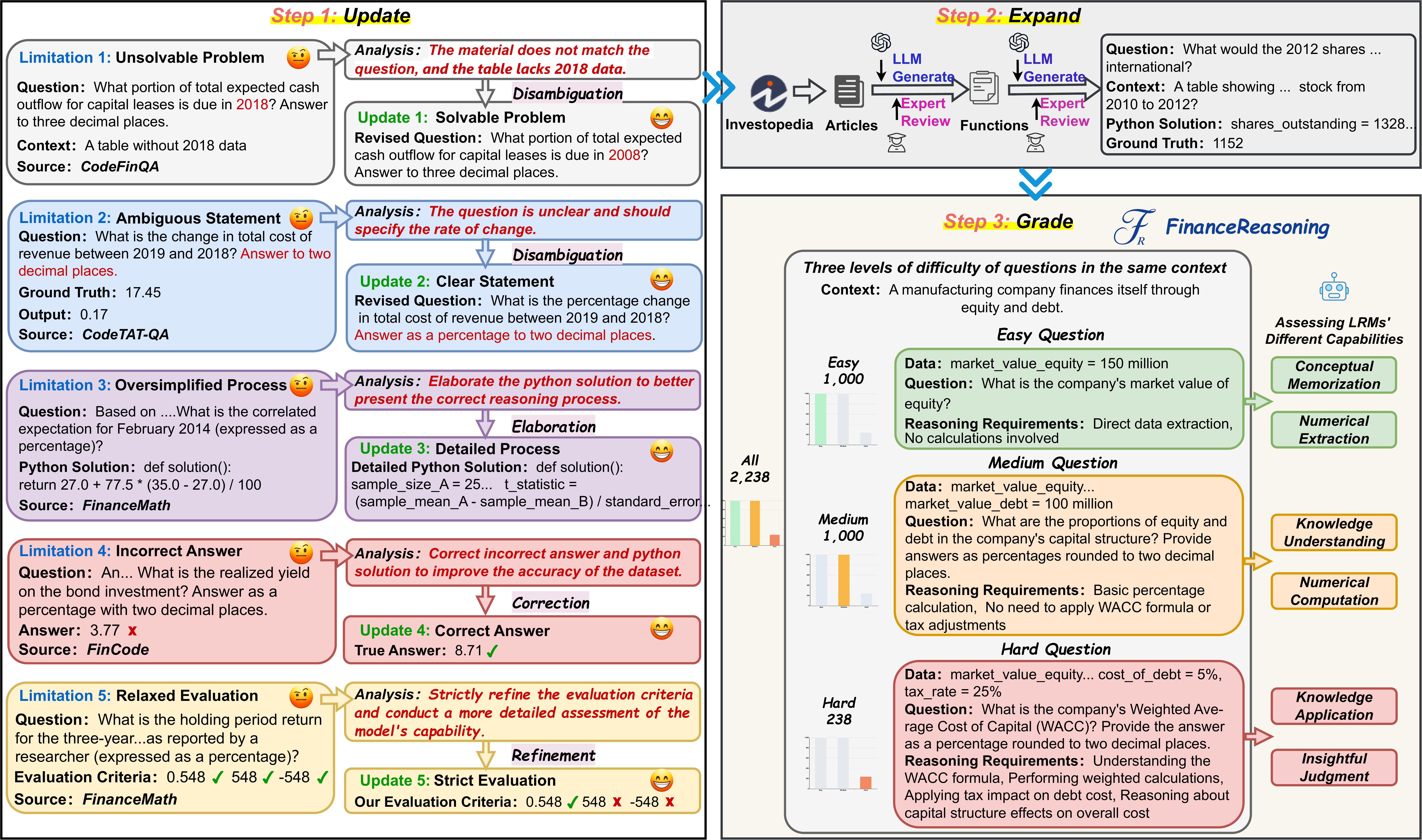}
    \caption{Overview of FinanceReasoning's construction.}
    \label{fig:overreview_v3}
\end{figure*}

% FinanceReasoning comprises 2,238 problems, with 1,420 problems refined based on existing datasets, while 908 problems were automatically generated by LLMs utilizing our open-sourced financial function library containing 3,133 Python-formatted functions. Each problem is accompanied by Python-formatted solutions and precise answers, with the entire process meticulously verified by experts. FinanceReasoning is categorized into three tiers of difficulty in terms of computational and logical complexity, each designed to evaluate distinct model capabilities.
Our experimental results demonstrate that the powerful LRM (\ie{ OpenAI o1}) with PoT achieves the best performance, with an accuracy of 89.1\% on \emph{Hard} subset, significantly outperforming other LLMs. However, current LRMs still faced incorrect formula application and imprecise numerical calculation on challenging domain-specific reasoning problems.
Next, we explore various knowledge augmentation methods and combinations of models. Experiments demonstrate that integrating structured, refined reasoning knowledge and enabling model collaboration can further enhance the complex reasoning capabilities of LRMs.

Our contributions are summarized below:
\begin{itemize} [leftmargin=*]
\itemsep0em 
\item We propose FinanceReasoning, a credible financial numerical reasoning benchmark constructed from re-annotated public datasets and newly collected challenging data through Human-AI collaboration, demonstrating the superior reasoning capabilities of LRMs.
\item We construct and open-source a comprehensive financial function library containing 3,133 Python-formatted functions, demonstrating the effectiveness of refined knowledge augmentation in enhancing domain-specific reasoning.
\item We analyze the shortcomings of LRMs and propose a combination of \textbf{Reasoner} and \textbf{Programmer} models, effectively enhancing their performance on complex mathematical calculations.
\end{itemize}

\section{\ours Benchmark}

As illustrated in \autoref{fig:overreview_v3}, we first update existing datasets like BizBench~\cite{krumdick-etal-2024-bizbench} and FinanceMath~\cite{zhao-etal-2024-knowledgefmath}, addressing issues such as disambiguation and corrections. We then construct a financial function library by extracting articles. Expert annotators are guided to review and revise the model-generated problems.

\subsection{Updates to Public Datasets}

Following prior work~\cite{krumdick-etal-2024-bizbench, zhao-etal-2024-knowledgefmath}, we retain the format of questions with optional hybrid contexts as input, accompanied by Python-formatted solutions and program-executed numerical results. Due to the specialized nature of complex financial problems and the high cost of expert annotation, we observe certain limitations in existing datasets~\cite{chen-etal-2021-finqa, zhu-etal-2021-tat, zhao-etal-2024-knowledgefmath}, including unsolvable problems, ambiguous statements (\eg{ the phrasing ``the range of'' confuses the LLMs, as it is not clear whether to output a range like 70.18-81.05 or a specific difference of 10.87}), oversimplified processes, incorrect answers, and relaxed evaluation criteria. Statistical details of these issues are presented in \autoref{tab:Table1}.

Specifically, we perform updates on the \emph{test} sets of CodeFinQA, CodeTAT-QA, and FinCode~\cite{krumdick-etal-2024-bizbench}, as well as the \emph{validation} set of FinanceMath~\cite{zhao-etal-2024-knowledgefmath}. The annotators are instructed to examine each example and perform three types of \textbf{Update Actions}: \textbf{Disambiguation}, \textbf{Elaboration}, and \textbf{Correction}.
\begin{itemize} [leftmargin=*]
\itemsep0em 
\item \textbf{Disambiguation}: For problems that are unsolvable due to insufficient contextual conditions or unclear target results, minimally modifies the question to eliminate potential ambiguities.

\item \textbf{Elaboration}: For Python programs with missing or skipped computational steps, supplement the code and add detailed annotations.

\item \textbf{Correction}: For problems with incorrect ground truth, the solution and answer are revised.
\end{itemize}

Additionally, existing evaluation standards are relatively relaxed: BizBench allows an error margin 1\%~\cite{krumdick-etal-2024-bizbench}, while FinanceMath disregards units and signs~\cite{zhao-etal-2024-knowledgefmath}. We refine the evaluation criteria by specifying units, percentage formats, signs, and decimal places, and strictly enforce a 0.2\% error margin, enhancing the rigor, challenge, and relevance to real-world scenarios. Detailed examples are illustrated in Appendix \ref{app:revise_example}.

\subsection{Function Library Construction}
 For LRMs, the challenge lies not in extracting numerical values from relevant texts but in applying domain-specific knowledge to perform complex multi-step numerical computations~\cite{Plaat2024ReasoningWL,chen-etal-2023-theoremqa}. Although LLMs have already acquired a solid understanding of conceptual knowledge in the financial domain, to further refine reasoning capabilities, we collect and annotate a financial function library comprising \nfunction structured Python functions for financial calculations, aimed at improving models' reasoning knowledge.

\begin{figure}[!t]
    \centering
    \includegraphics[width = \linewidth]
    {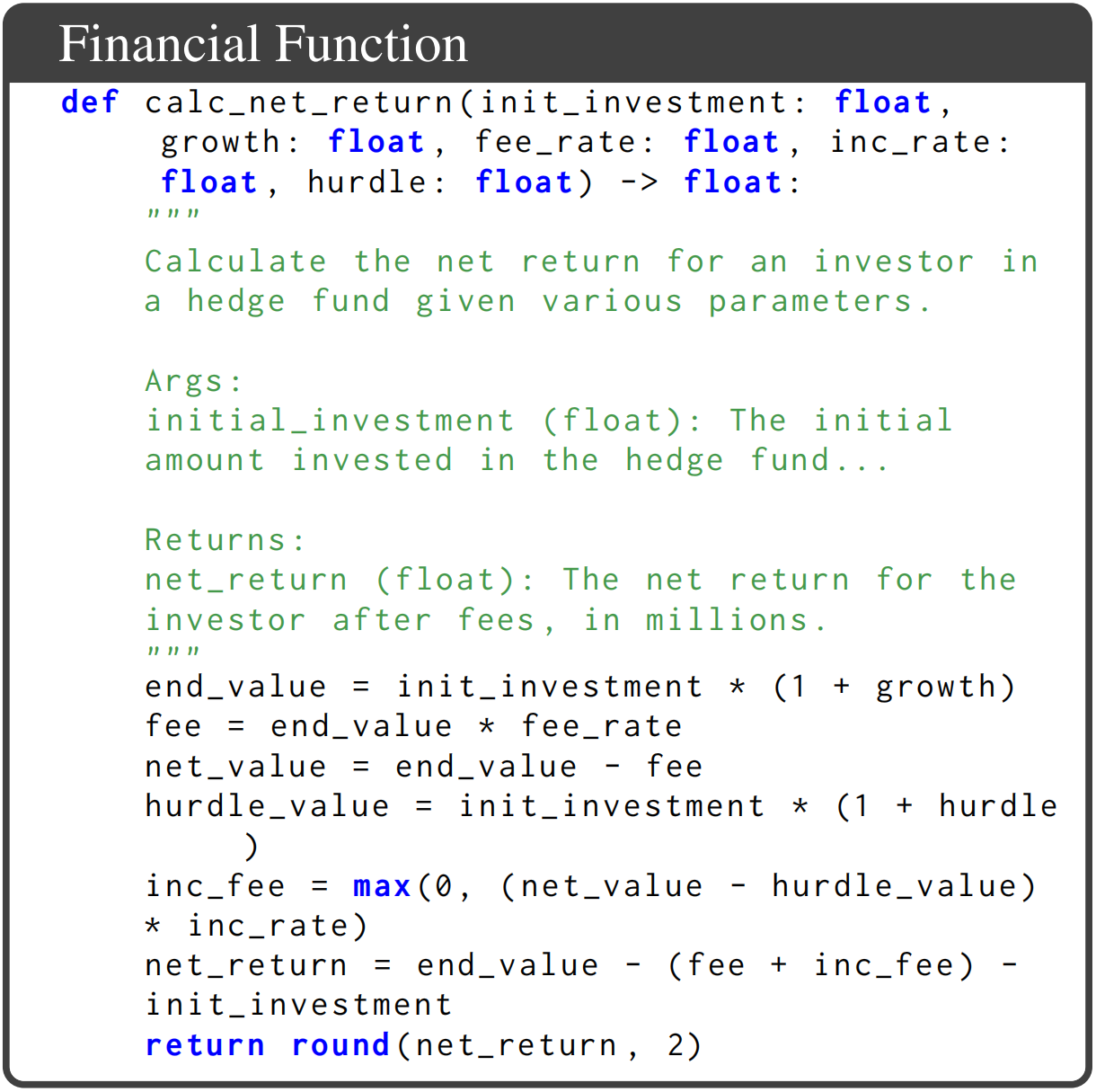}
    \caption{An example of financial function from the constructed function library to calculate net return.}
    \label{fig:financial-function}
\end{figure}

 We begin by collecting \narticle financial encyclopedia articles from Investopedia, a platform renowned for its extensive expertise in financial knowledge\footnote{\url{https://www.investopedia.com}}. Each article provides a detailed introduction to a specific financial term, covering fundamental concepts, application scenarios, and potential limitations, some including relevant calculation formulas and practical examples. To distill dense, structured financial reasoning knowledge while reducing annotation costs, we instruct GPT-4o to extract potential financial calculation functions from each article according to a predefined format. Each function is required to include a semantically meaningful signature, a concise and clear docstring (functionality, parameters, return values, applicable constraints, and other notes), and step-by-step implementation code with appropriate annotations. Finally, we organize financial experts to rigorously review and revise the generated functions, ensuring their professional expression and logical correctness. Detailed examples of financial functions are illustrated in \autoref{fig:financial-function} and \autoref{tab:exmFunc}.

 \subsection{Expansion of Data Annotation}
Existing financial question-answering datasets (\eg CodeFinQA, CodeTAT-QA~\cite{krumdick-etal-2024-bizbench}) focus primarily on evaluating models' basic concept understanding, precise numerical extraction, and simple calculation abilities within given contexts. Problem-solving processes in these datasets typically involve fewer reasoning steps (\eg calculate the difference in net profit over two years). These datasets often suffer from redundancy in simple questions and a lack of complex questions, failing to adequately assess the reasoning capabilities of LRMs, such as knowledge application, constraint emphasis and long thought (\eg compute the net return rate of a fund in \autoref{fig:example}). As a result, the real reasoning capabilities of LRMs cannot be evaluated comprehensively and objectively. Therefore, optimizing data construction methods, rigorously verifying data quality, and building more challenging datasets have become crucial to improve the evaluation of financial reasoning tasks.

During the data expansion process, we leveraged the structured financial functions to guide GPT-4o in generating new financial numerical reasoning problems and Python solutions. Then, experts rigorously reviewed and corrected them, resulting in 908 high-quality problems with varying reasoning difficulties and a wide knowledge coverage. The data annotation process is as follows:

\paragraph{Seed Function Selection}
We selected 1,250 financial functions from the library based on operators, arguments, code lines, and long-tail knowledge, prioritizing those with complex computation.

\paragraph{Question and Solution Generation}
For each seed function, GPT-4o was prompted to generate the complex reasoning problem with the necessary financial tabular data, using the financial terms and the computational processes of the function. The generated Python solutions were required to have clear reasoning paths and be executable to acquire numerical answers, taking into account units, percentages, and decimal precision requirements.

\paragraph{Expert Verification}
The experts are required to review and correct all problems, solutions, and answers to ensure the absence of ambiguities, detailed processes, and correct answers.

\subsection{Data Quality Assurance}
To ensure the high quality of \ours, we implemented a rigorous annotation process. Specifically, we organized a team of 8 graduate students with interdisciplinary backgrounds in finance and computer science, along with 2 experts holding CFA licenses, to participate in the dataset verification. Each financial function and problem were initially reviewed by two graduate students, who provided reasons for errors and suggested modifications. Consistent suggestions were adopted directly. For cases with conflicting opinions, the final modification plan was determined through a discussion between the two experts. With the help of LLMs, the entire annotation process lasted for three months. The annotation example is provided in Appendix \ref{app:annotation_case}.
\begin{table}[t!]
\centering
\resizebox{\columnwidth}{!}{
\addtolength{\tabcolsep}{-0.5em}
\begin{tabular}{lr}
\toprule
\textbf{Property}                            & \textbf{Value}  \\
\midrule
\multicolumn{2}{c}{\textbf{Financial Function Library}} \\
\noalign{\vskip 1ex}
\# Total Functions & 3,313 \\
\noalign{\vskip 0.5ex}\hdashline\noalign{\vskip 0.5ex}
\# Operators (Avg) & 2.85 \\
\# Arguments (Avg) & 2.64 \\
\# Lines of Code (Avg) & 3.45 \\
\# Financial Concepts Involved & 1,864 \\
\midrule

\multicolumn{2}{c}{\textbf{FinanceReasoning Dataset}} \\
\noalign{\vskip 1ex}
\# Operators (Easy/Medium/Hard) & 1.77/3.79/\textbf{10.12} \\
\# Lines of Code (Easy/Medium/Hard) & 3.13/4.27/\textbf{9.49}\\
\# Parentheses (Easy/Medium/Hard) & 0.80/3.28/\textbf{11.21} \\
\# Difficulty (Easy/Medium/Hard) & 1.69/3.00/\textbf{4.88} \\
\bottomrule
\end{tabular}
}
\caption{Statistics of the financial function library and FinanceReasoning dataset (Avg values of three subsets).}
\label{tab:statistics}
\end{table}

\begin{figure}[!t]
    \centering
    \includegraphics[width = 1\linewidth]{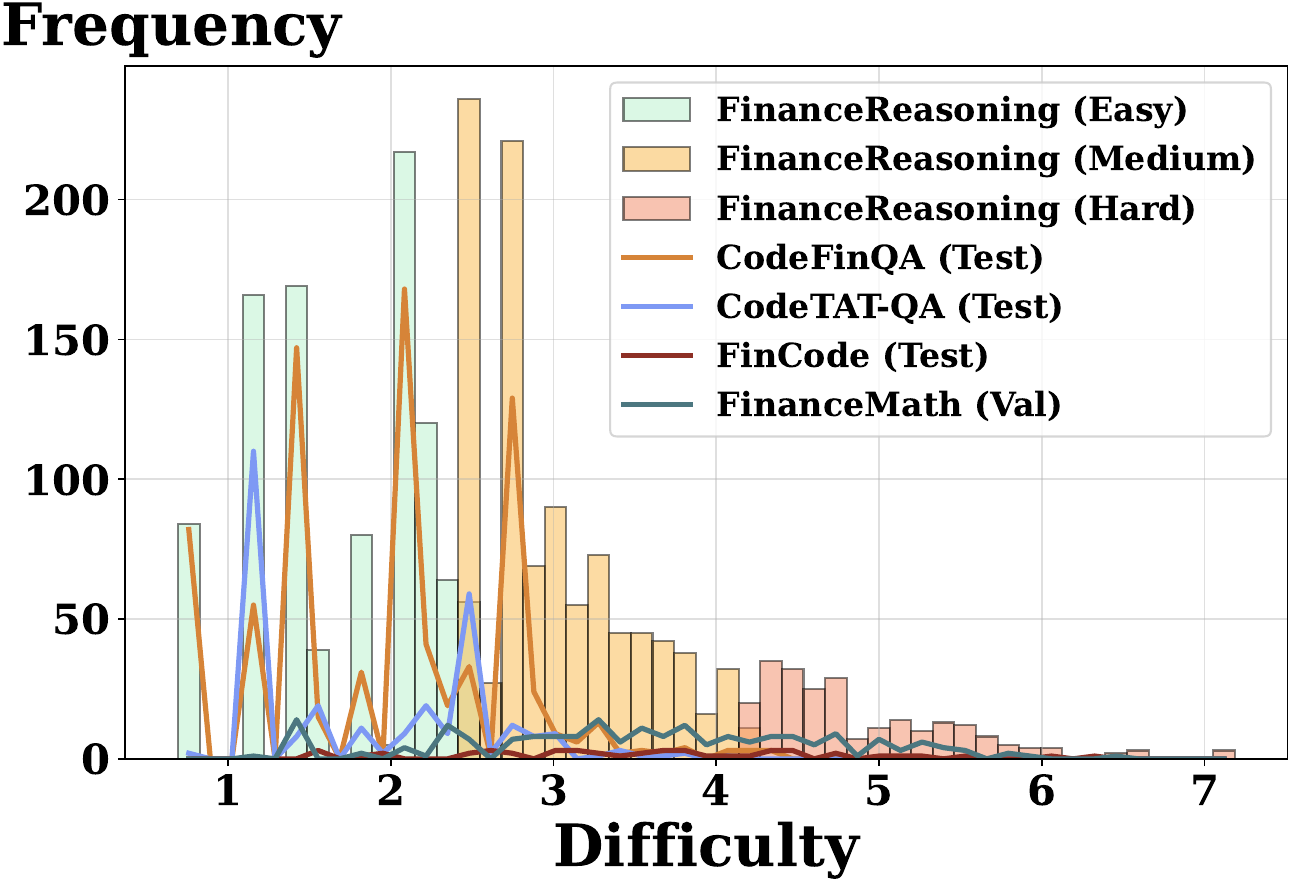}
    \caption{The difficulty distribution of \ours, compared with four existing datasets, shows a notably higher proportion of medium and hard problems, presenting greater challenges for complex reasoning.}
    \label{fig:difficulty_distribution}
\end{figure}
\subsection{Data Grading and Statistics}
To evaluate the performance of LRMs in financial numerical reasoning problems of varying difficulty levels, we designed a heuristic algorithm for the first time to assess the difficulty of reasoning for each problem based on the number of operators, pairs of parentheses, and lines of code in the Python program. Specifically, the difficulty of reasoning \( rc \) of a problem is defined as:
\begin{equation}
rc = \ln \left( \max \left( o, 1 \right) \right) + \ln \left( \max \left( l + p, 1 \right) \right)
\label{eq:difficulty}
\end{equation}
where \( o \) is the number of operators, \( p \) is the number of pairs of parentheses, and \( l \) is the number of code lines in the Python program. 

As illustrated in \autoref{tab:statistics} and \autoref{fig:difficulty_distribution}, based on the difficulty of reasoning, we divided the problems into three subsets: \emph{Easy} (1,000 examples), \emph{Medium} (1,000 examples), and \emph{Hard} (238 examples). More analyses are in Appendix  \ref{app:difficulty}. To promote the evaluation of LRMs' reasoning capabilities in the financial domain, we have made \textbf{all problems publicly available}, along with the \textbf{complete financial function library}.
\begin{table*}[!ht]

\centering
\fontsize{8pt}{10pt}\selectfont
\resizebox{\textwidth}{!}{%
\renewcommand{\arraystretch}{1.0}
\addtolength{\tabcolsep}{-0.1em}
\begin{tabular}{lrlrrrrrrrrrrr}
\toprule
\multirow{2}{*}{\textbf{Model}} & \multirow{2}{*}{\textbf{Size}}  & \multirow{2}{*}{\textbf{Notes}} & \multicolumn{2}{c}{\textbf{Hard}} & ~ & \multicolumn{2}{c}{\textbf{Medium}} & ~ & \multicolumn{2}{c}{\textbf{Easy}} & ~ & \multicolumn{2}{c}{\textbf{Avg.}}\\
\cmidrule(lr){4-5} \cmidrule(lr){7-8} \cmidrule(lr){10-11} \cmidrule(lr){13-14}
~ & ~ & ~ & CoT & PoT & & CoT & PoT & & CoT & PoT & & CoT & PoT\\
\midrule
\multicolumn{14}{l}{\emph{\textbf{Large Reasoning Models (LRMs)}}} \\
OpenAI o1 & ~ & ~ & \cellcolor{yellow!25} 81.1 & \cellcolor{yellow!45} \textbf{89.1} & & \cellcolor{yellow!25} 89.7 & -- & & 88.0 & -- & & \cellcolor{yellow!25} 86.3 & -- \\
DeepSeek-R1 & 671B & MoE & \cellcolor{yellow!45} \textbf{83.2} & \cellcolor{yellow!25} 85.3 & & \cellcolor{yellow!45} \textbf{91.1} & \cellcolor{yellow!45} \textbf{89.8} & & \cellcolor{yellow!45} \textbf{89.8} & \cellcolor{yellow!45} \textbf{89.2} & & \cellcolor{yellow!45} \textbf{88.0} & \cellcolor{yellow!45} \textbf{88.1}\\
OpenAI o3-mini & ~ & ~ & \cellcolor{yellow!15} 77.3 & \cellcolor{yellow!15} 84.0 & & 87.8 & \cellcolor{yellow!25} 88.6 & & \cellcolor{yellow!15} 88.8 & 88.1 & & \cellcolor{yellow!15} 84.6 & 86.9 \\ 
OpenAI o1-mini & ~ & ~ & 71.4 & 83.6 & & 86.2 & 86.9 & & 85.6 & 87.0 & & 81.1 & 85.8 \\ 
Gemini 2.0 Flash Thinking Experimental & ~ & ~ & 70.6 & 81.5 & & 85.2 & 87.2 & & \cellcolor{yellow!15} 88.8 & 86.6 & & 81.5 & 85.1\\ 
QwQ-32B-Preview & 32B & ~ & 63.5 & 61.8 & & 81.1 & 72.8 & & 83.5 & 74.9 & & 76.0 & 69.8\\ 
\midrule

\multicolumn{14}{l}{\emph{\textbf{Large Language Models (LLMs)}}} \\
Gemini 2.0 Pro Experimental & ~ & ~ & 72.3 & 83.6 & & \cellcolor{yellow!15} 88.3 & 87.4 & & 87.3 & 87.8 & & 82.6 & 86.3\\
Claude 3.5 Sonnet & ~ & ~ & 68.5 & 83.6 & & 85.7 & \cellcolor{yellow!15} 88.2 & & 87.7 & \cellcolor{yellow!15} 88.4 & & 80.6 & \cellcolor{yellow!25} 86.7\\
GPT-4o & ~ & ~ & 65.6 & 83.6 & & 84.6 & 87.9 & & 86.8 & 88.1 & & 79.0 & \cellcolor{yellow!15} 86.5\\
Qwen2.5-Max & ~ & MoE & 65.1 & 82.4 & & 87.2 & 86.5 & & \cellcolor{yellow!25} 89.6 & \cellcolor{yellow!25} 89.1 & & 80.6 & 86.0\\
DeepSeek-V3 & 671B & MoE & 66.8 & 75.6 & & 85.2 & 87.3 & & 87.2 & 86.9 & & 79.7 & 80.7\\ 
Llama 3.3 & 70B & ~ & 50.4 & 71.4 & & 79.2 & 85.9 & & 83.3 & 84.8 & & 71.0 & 80.7\\
Llama 3.1 & 405B & ~ & 51.7 & 70.2 & & 81.7 & 87.7 & & 84.1 & 85.8 & & 72.5 & 81.2\\ 
    \bottomrule
    \end{tabular}
    }
\caption{Results of different models using CoT and PoT prompting methods on the different subsets of FinanceReasoning. We use Accuracy of \emph{Hard} subset using PoT prompting as the ranking indicator of model performance. The results underscore the superior performance of LRMs (\ie{ OpenAI o1 and DeepSeek-R1}) with PoT.}
\label{tab:main_result}
\end{table*}

% Performance comparison of different models using CoT and PoT prompting methods on the FinanceReasoning benchmark. This table presents a comparative analysis of Large Reasoning Models (LRMs) and Large Language Models (LLMs) across varying levels of reasoning complexity. The results underscore the efficacy of CoT in enhancing logical reasoning within LRMs, while PoT demonstrates superior computational efficiency in structured problem-solving for LLMs.

\section{Evaluation System}
We developed an evaluation system for complex reasoning on FinanceReasoning, where all evaluations of LLMs were conducted by calling their official API interfaces. \autoref{tab:model-detail} illustrates the exact model versions we used.

\subsection{Large Language Models}
We focused on evaluating six of the most powerful LRMs currently available.
\begin{itemize} [leftmargin=*]
\itemsep0em 
\item \textbf{OpenAI o1}~\cite{openai2024o1} is trained using large-scale reinforcement learning and employs chain-of-thought reasoning, excelling in general knowledge tasks and code reasoning tasks.

\item \textbf{OpenAI o1-mini}~\cite{openai2024o1mini} is a cost-effective alternative to OpenAI o1, designed for high performance in STEM fields, particularly in mathematics and coding.

\item \textbf{OpenAI o3-mini}~\cite{openai2025o3mini} is OpenAI's latest small reasoning model, providing faster response times while maintaining comparable performance to OpenAI o1.

\item \textbf{DeepSeek-R1}~\cite{guo2025deepseekr1} enhances its reasoning capabilities through multi-stage training with reinforcement learning, using a minimal amount of supervised fine-tuning data. 

\item \textbf{Gemini 2.0 Flash Thinking Experimental}~\cite{geminiteam2025gemini} introduces a 1M token context window to deeply understand long texts and incorporates self-correction mechanisms.
    
\item \textbf{QwQ-32B-Preview}~\cite{qwq-32b-preview} is an experimental model of the Qwen team, approaching problems with curiosity, self-questioning, and reflection, striving for a deeper understanding. 
 
\end{itemize}

Gemini 2.0 Pro Experimental~\cite{geminiteam2025gemini}, GPT-4o~\cite{openai2024gpt4o}, Claude 3.5 Sonnet~
\cite{claude3.5sonnet}, DeepSeek-V3~
\cite{deepseekai2024deepseekv3}, Llama 3.3~
\cite{llama3.3modelcard}, Llama 3.1~
\cite{llama3.1modelcard}, and Qwen2.5-Max~
\cite{qwen2025qwen25} are also evaluated for comparison, providing a baseline to assess the performance between LRMs and traditional LLMs.

\subsection{Evaluation Methods}
\paragraph{Prompting Methods}
Following \citet{zhao-etal-2024-knowledgefmath}, we evaluated LLMs with CoT~\cite{wei2022chain} and PoT~\cite{chen2023program} to achieve optimal performance and make comparisons.

\paragraph{Answer Extraction and Evaluation}
We adopt the answer extraction pipeline from \citet{zhao-etal-2024-knowledgefmath}, using GPT-4o-mini to extract numerical results from the output in CoT setting, and executing the program from the output in PoT setting. Finally, we perform a strict accuracy evaluation comparing the numerical results with the ground truth within 0.2\% error margin.

\section{Experiments}
We answer the following research questions (RQs):
\textbf{RQ1}: Do LRMs outperform other LLMs in financial reasoning tasks?
\textbf{RQ2}: What are the main shortcomings of LRMs?
\textbf{RQ3}: Does refined knowledge augmentation improve LRMs' performance?
\textbf{RQ4}: Does model collaboration enhance LRMs' performance?
\textbf{RQ5}: Does PoT outperform CoT in complex numerical reasoning tasks?

\subsection{Main Results (RQ1)}
The performance of the evaluated LRMs and LLMs using two prompting methods on the FinanceReasoning are shown in \autoref{tab:main_result}. 

The results demonstrate that the powerful LRM (\ie{ OpenAI o1}) using PoT prompting method achieves the best performance, with an accuracy of 89.1\% on the \emph{Hard} subset, significantly outperforming other LRMs and LLMs. On the \emph{Easy} and \emph{Medium} subsets, the evaluated LLMs achieve accuracy above 87\%, except for the Llama models, where the advantage of LRMs is less pronounced. This further validates that simpler datasets have largely been solved by LLMs, making it difficult to assess the real reasoning capabilities of LRMs. On the \emph{Hard} subset, LRMs exhibit a clear advantage over LLMs with CoT, further confirming the superiority of LRMs in complex reasoning tasks.

In particular, on the \emph{Hard} subset, the current superior LRMs (\ie{ OpenAI o1 and DeepSeek-R1}) exhibit distinct performance contrasts in CoT and PoT settings. In CoT, DeepSeek-R1 achieves a 2.1\% higher accuracy than OpenAI o1 \textbf{(83.2\% vs. 81.1\%)}. However, the PoT prompting method significantly enhances OpenAI o1's performance, allowing it to surpass DeepSeek-R1 \textbf{(89.1\% vs. 85.3\%)}. This suggests that DeepSeek-R1 excels at text-based step-by-step reasoning, while OpenAI o1 outperforms in programming capabilities. This discrepancy may be due to differences in their training methods and training data.

\subsection{Error Analysis (RQ2)}
To better analyze the capabilities and limitations of LRMs on difficult problems in our dataset, we conduct a thorough and comprehensive error analysis. This analysis is based on 80 DeepSeek-R1 failure cases with PoT, with stratified sampling (20 \emph{Easy} /20 \emph{Medium} /40 \emph{Hard}). We summarize four types of error in the current LRMs on challenging domain-specific reasoning problems, some of which involve compound errors. The detailed error distribution is shown in \autoref{tab:error-distribution}. More details of error cases are provided in Appendix \ref{app:error_case}.
\begin{itemize} [leftmargin=*]
\itemsep0em
\item \textbf{Misunderstanding of Problem}: The model incorrectly interprets the question and context due to a lack of financial knowledge.

\item \textbf{Formula Application Errors}: Owing to inexperience in financial reasoning, the model uses an incorrect formula that does not correspond to the specified conditions of the problem.

\item \textbf{Numerical Extraction Errors}: The model extracts incorrect variables, especially when processing structured tabular data, despite the fact that the reasoning process and the selected formula are correct.

\item \textbf{Numerical Calculation Errors}: When multiple calculation steps are involved, the model produces significant precision differences from the correct answer due to rounding and hallucination during the computation process.
\end{itemize}

\begin{table}[tbp]
  \centering
  \footnotesize
  \begin{tabularx}{\linewidth}{@{}>{\centering\arraybackslash}X*{4}{>{\centering\arraybackslash}X}@{}}
    \toprule
    \textbf{Subset} & \textbf{M} & \textbf{F} & \textbf{E} & \textbf{C} \\
    \midrule
    Easy   & \cellcolor{yellow!25}35\% (7/20)  & 5\% (1/20)   & \cellcolor{yellow!15}15\% (3/20) & \cellcolor{yellow!45}45\% (9/20)  \\
    Medium & \cellcolor{yellow!15}25\% (5/20)  & \cellcolor{yellow!25}30\% (6/20)  & 5\% (1/20)  & \cellcolor{yellow!45}40\% (8/20)  \\
    Hard   & \cellcolor{yellow!15}20\% (8/40)  & \cellcolor{yellow!25}35\% (14/40) & 7.5\% (3/40)& \cellcolor{yellow!45}37.5\% (15/40)\\
    \bottomrule
  \end{tabularx}
  \caption{\footnotesize Error distribution across difficulty subsets. M means Misunderstanding of Problem, F means Formula Application Errors, E means Numerical Extraction Errors, C means Numerical Calculation Errors.}
  \label{tab:error-distribution}
\end{table}

As shown in \autoref{tab:error-distribution}, we observe that:
\begin{itemize} [leftmargin=*]
\itemsep0em
\item \textbf{Formula Application Errors} and \textbf{Numerical Calculation Errors} increase with difficulty, especially the former, which supports the effectiveness of our difficulty categorization. In contrast, \textbf{Numerical Extraction Errors} remain consistently low across all subsets. These findings suggest that the model's domain-specific knowledge comprehension and application capabilities should be enhanced through retrieval-augmented generation (RAG) or reinforcement learning (RL), thereby improving its performance on complex reasoning tasks in specialized domains.

\item \textbf{Numerical Calculation Errors} occur at similar rates across all difficulty levels. Since \textbf{Formula Application Errors} are less frequent in easier subsets, improving the handling of calculation-related issues, such as unit conversions, percentage formats, significant digits, and sign correctness, becomes key to boosting accuracy on \emph{Easy} and \emph{Medium} problems. The results indicate the necessity of incorporating external computational tools or model collaboration to strengthen numerical computation capabilities, which could mitigate errors caused by calculation inaccuracies on relatively simple problems.
\end{itemize}

\subsection{Knowledge Augmentation (RQ3)}
To enhance the understanding and application capabilities of complex formulas of LLMs in financial reasoning tasks, we explored and compared two formats of knowledge and various methods of enhancing knowledge to improve the performance of LLMs in domain-specific reasoning tasks.

\paragraph{Knowledge Augmentation Settings}
We use Contriever~\cite{izacardunsupervised} to retrieve relevant financial knowledge passages or financial Python functions based on the question.
\begin{itemize} [leftmargin=*]
\itemsep0em 
\item \textbf{Function Retrieval}: We use the question as a query to compute semantic similarity with the function descriptions, retrieving the Top-3 financial functions as relevant knowledge.
    
\item \textbf{Passage Retrieval}: For comparison with function retrieval, we segment each collected financial article into passages based on markdown hierarchical structures and retrieved the Top-10 passages for knowledge enhancement.
    
\item \textbf{LLM as Retrieval Judge}: Recent studies have shown that models are capable of judging the relevance of candidates retrieved for the question~\cite{guan2024amor}. In this setting, we first retrieved the Top-30 financial functions and then prompted the LLM to select the Top-3 functions most useful to answer the question, if any.
    
\item \textbf{LLM-Instructed Knowledge Retrieval}: In financial problems with hybrid contexts, using short questions or full contexts for retrieval often fails to retrieve directly relevant knowledge~\cite{chen-etal-2023-beyond,peng2023check}. We observed that powerful LLMs (\eg GPT-4o) can effectively summarize rich semantic information from contexts. Therefore, we prompt the LLM to generate precise retrieval queries based on the context~\cite{li2025searcho1,verma2025plan}.

\end{itemize}

\begin{table}[!t]
    \centering
    \resizebox{\linewidth}{!}{
    \begin{tabular}{lll}
        \toprule
        Setting & GPT-4o (PoT) & DeepSeek-R1 (CoT)\\
        \midrule 
        \emph{wo.} knowledge augmentation & 83.19 & 83.19\\ 
        \noalign{\vskip 0.5ex}\hdashline

        \noalign{\vskip 0.5ex}
        Passage Retrieval ($n=10$)& \\
        \quad Vanilla Retrieval & 81.93 \down{1.26} & 82.77 \down{0.42}\\
        
        \noalign{\vskip 0.5ex}\hdashline\noalign{\vskip 0.5ex}
        Function Retrieval ($n=3$)& \\
        \quad Vanilla Retrieval & 90.76 \up{7.57} & 85.29 \up{2.10}\\
        \quad LLM as Judge & 89.08 \up{5.89} & 84.87 \up{1.68}\\
        \quad \makecell[l]{LLM-instructed Retrieval  \&\\ LLM as Judge} & \textbf{91.60} \up{8.41} & \textbf{86.97} \up{3.78}\\
        
        \bottomrule
    \end{tabular}
    }
    \caption{Results of different knowledge augmentation methods on the \emph{Hard} subset of \ours. GPT-4o with refined knowledge augmentation, outperforming OpenAI o1 \textbf{(91.6\% vs. 89.1\%)} in PoT setting.}
    \label{tab:retriever_performance}
\end{table}
\paragraph{Knowledge Augmentation Results}
As shown in \autoref{tab:retriever_performance}, the format and method of knowledge augmentation significantly affect the performance of the model reasoning. Specifically, LLMs enhanced with financial function knowledge significantly outperform those enhanced with passage knowledge, as financial functions serve as refined reasoning knowledge. Excessive and intricate passages can disrupt the model's reasoning abilities, resulting in diminished performance for both LLMs and LRMs. Taking advantage of the improved retrieval efficiency caused by \textbf{LLM-Instructed Knowledge Retrieval}, the combination approach achieves the best performance, improving the accuracy of GPT-4o to 91.6\% with PoT. More analyses are shown in Appendix \ref{app:rag_case}.

\subsection{Reasoner with Programmer (RQ4)}
To address the issue of imprecise numerical calculations in LRMs, we instruct the LRM to act as the \textbf{Reasoner}, responsible for carefully reasoning through the problem-solving path, while disregarding its generated numerical results. Then, a code-specialized LLM acts as the \textbf{Programmer}, strictly following the reasoning path provided by the Reasoner to generate executable Python programs, which are ultimately executed to obtain precise numerical results.
\begin{figure}[!t]
    \centering
    \includegraphics[width = \linewidth]{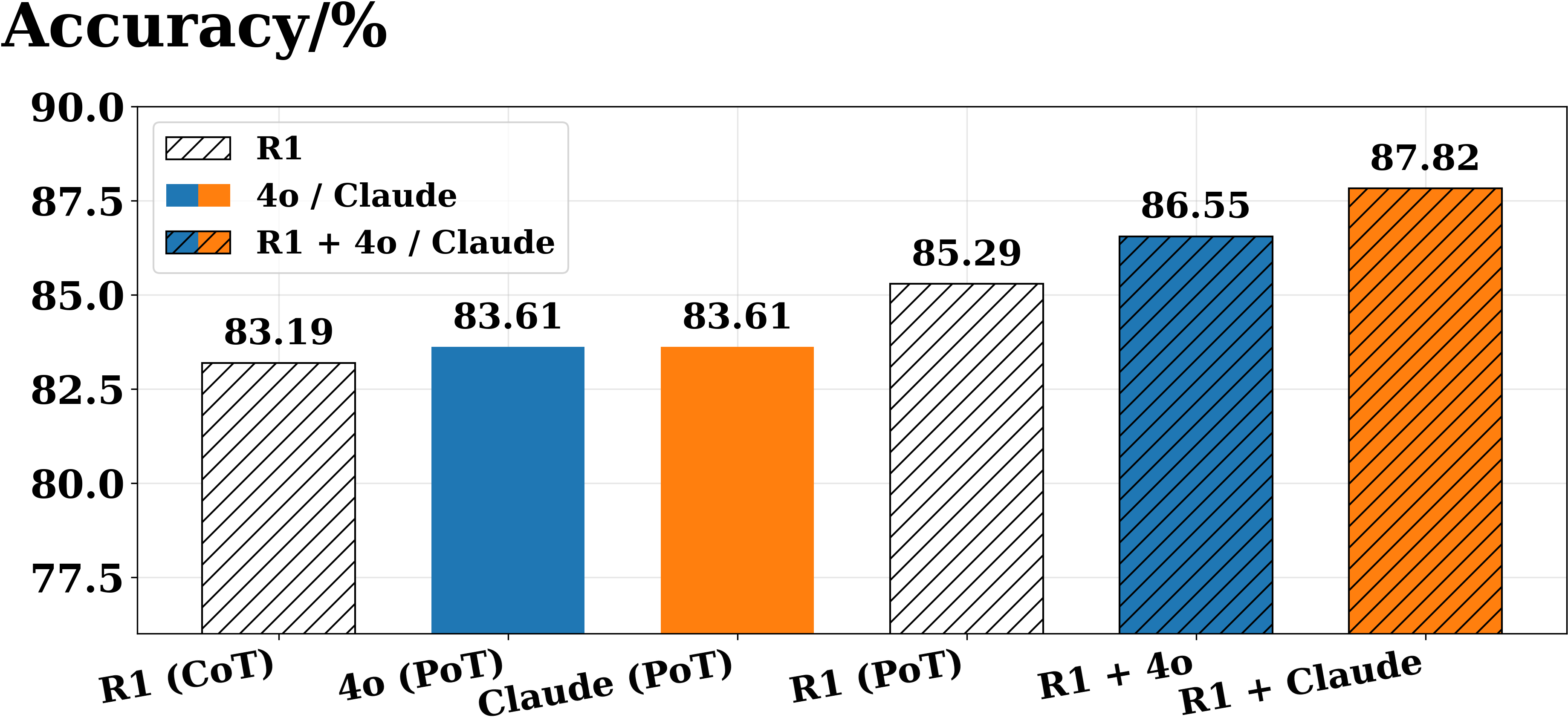}
    \caption{Results of different model combinations and individual models. R1 means DeepSeek-R1, 4o means GPT-4o, Claude means Claude 3.5 Sonnet.}
    \label{fig:rag}
\end{figure}

Specifically, we employ \textbf{DeepSeek-R1} as Reasoner, and the best-performing models in PoT setting, \textbf{Claude 3.5 Sonnet} and \textbf{GPT-4o}, as Programmers, respectively. As shown in \autoref{fig:rag}, the combination of models achieves significant improvements compared to individual models. Compared to GPT-4o, Claude 3.5 Sonnet demonstrates a stronger ability to follow the given reasoning logic and generate precise code without introducing noise, owing to its programming advantages. The combination of DeepSeek-R1 and Claude 3.5 Sonnet achieves an accuracy of 87.82\%, correcting 91.7\% of the numerical calculation errors. Furthermore, we isolate the knowledge reasoning capabilities of DeepSeek-R1 in complex financial reasoning tasks, which outperform other LLMs. More analyses are shown in Appendix \ref{app:reasoner_cot_case}.

\subsection{PoT vs.\ CoT (RQ5)}  
Based on an analysis of the \emph{Hard} subset, we observe that PoT exhibits a markedly stronger performance than CoT in multi-step and complex numerical reasoning tasks. Specifically, PoT leverages structured code generation to reduce token consumption. Under similar or lower token usage, PoT achieves greater accuracy (\autoref{tab:main_result}). Moreover, its performance is on par with certain LRMs that utilize test-time scaling strategies. For example, GPT-4o, when prompted with PoT, consumes only 54k tokens to solve the \emph{Hard} subset, whereas CoT requires 173k tokens. Detailed statistics on token consumption are provided in \autoref{tab:tokens}.

As shown in \autoref{fig:cost_effective}, LLMs using PoT prompting method not only significantly reduce token overhead during inference, but can also approach or match the performance of LRMs. DeepSeek-R1 with CoT achieves the similar accuracy on the \emph{Hard} subset as GPT-4o with PoT, but consumes much more tokens \textbf{(742k vs. 54k)}. This ``token-for-accuracy'' test-time scaling strategy allows LRMs to maintain high correctness by repeatedly verifying outcomes from multiple perspectives. However, the associated inference cost is prohibitively high. In contrast, PoT achieves performance comparable to LRMs for complex financial calculations, while offering notably better cost-effectiveness.
\section{Related Work}

The emergence of reasoning models such as OpenAI o1~\cite{openai2024o1} and DeepSeek-R1~\cite{guo2025deepseekr1} has significantly improved the performance of LLMs in complex reasoning tasks in domains such as code~\cite{jain2024livecodebench, chen2021humaneval}, math~\cite{mao-etal-2024-champ, lightman2023let}, and science~\cite{lu2023mathvista, yue2023mmmu, wang2024scibench}. Among these LRMs~\cite{openai2024o1,openai2024o1mini,openai2025o3mini,guo2025deepseekr1,qwq-32b-preview,team2025kimi,geminiteam2025gemini}, OpenAI o1 and DeepSeek-R1 have achieved competitive optimal performance. However, there currently exists a gap between the evaluation of LRMs and real-world domain-specific reasoning tasks, with a lack of evaluation and research on the model's ability to flexibly apply domain knowledge in complex multi-step numerical reasoning. In the financial domain, the difficulty of questions and the quality of annotations become key limitations in evaluating the real reasoning capabilities of LRMs. For example, CodeFinQA and CodeTAT-QA~\cite{krumdick-etal-2024-bizbench}, which are derived from the classic financial question-answering datasets FinQA~\cite{chen-etal-2021-finqa} and TAT-QA~\cite{zhu-etal-2021-tat}, rely on tabular data extraction and simple arithmetic operations that cannot accurately assess the improvements in reasoning ability of LRMs compared to LLMs. For datasets such as FinCode~\cite{krumdick-etal-2024-bizbench} and FinanceMath~\cite{zhao-etal-2024-knowledgefmath}, limited complex problems, ambiguous questions, and relaxed evaluation criteria hinder the accurate assessment.

\section{Conclusion}

This paper introduces FinanceReasoning, a credible, comprehensive, and challenging benchmark designed to evaluate the financial numerical reasoning capabilities of LRMs. We update existing numerical reasoning financial question answering datasets, rigorously refine evaluation standards, and explore methods to build complex reasoning datasets tailored for LRM evaluation. We comprehensively evaluated the six most advanced LRMs in subsets of varying difficulty levels. Compared to LLMs, we validated the leading performance of OpenAI o1 and DeepSeek-R1, while highlighting the need for further improvement in the precise numerical reasoning capabilities among LRMs. Our experiments on knowledge augmentation and the combination of models demonstrate that integrating structured, refined reasoning knowledge and enabling model collaboration can further enhance the complex reasoning performance of LRMs in expert domains.

\section*{Limitations}

In this work, we introduce FinanceReasoning, a benchmark dataset designed to evaluate and enhance LLMs in complex financial numerical reasoning tasks that require multi-step quantitative analysis, precise formula application, and hybrid contextual understanding. However, there are still some limitations: (1) We process tabular content as text, whereas in real-world scenarios, tables may also appear as images, requiring additional processing steps. In such cases, datasets such as MathVista \cite{lu2023mathvista} and MMMU \cite{yue2023mmmu}, which focus on reasoning over image-based questions, serve as valuable complements to our benchmark. We believe that incorporating elements from these datasets into FinanceReasoning could help bridge the gap between text-based and multimodal financial reasoning, enabling a more comprehensive assessment of LLMs' real-world applicability. (2) Due to limited resources, we do not conduct an evaluation of OpenAI o1 with PoT on the \emph{Easy} and \emph{Medium} subsets, as preliminary experiments suggest they are less challenging, and existing LLMs already demonstrate strong performance on these levels. (3) While we systematically verified and updated numerical answers and program solutions for multiple published datasets, we were unable to perform the same verification for the 1,000-problem \emph{test} subset of FinanceMath \cite{zhao-etal-2024-knowledgefmath}, as it does not publicly provide ground-truth references or Python solutions, limiting our ability to ensure consistency in result validation. We commit to incorporating additional FinanceMath \emph{test} subset should it become publicly available in the future, subject to the same rigorous validation process. (4) While FinanceReasoning provides comprehensive evaluation criteria, its long-term utility depends on adapting to evolving challenges. To address this, we commit to maintaining and periodically updating the benchmark through versioned releases on Hugging Face Datasets. (5) We clarify that FinanceReasoning evaluates \textit{perfect-information} scenarios where all solution prerequisites are explicitly provided, thus focusing on the upper bounds of LRMs' precision in deterministic calculations. This design intentionally distinguishes reasoning errors (\eg{ outputting -3.1 vs. 310,000 given clear constraints}) from ambiguity resolution capabilities. While the important direction of modeling LRMs' proactive clarification-seeking behavior under insufficient conditions falls outside this work's scope, we will pursue it in future research through controlled ambiguity injection and interactive evaluation protocols.
\section*{Ethical Considerations}
This work complies with ACL ethics guidelines. FinanceReasoning is released under the Creative Commons Attribution 4.0 International License (CC BY 4.0), permitting free use with proper attribution. The codebase is distributed under the Apache License 2.0, ensuring compatibility with commercial and open-source ecosystems.
\section*{Acknowledgements}
This work is supported by the National Natural Science Foundation of China (Grant Nos. 62176026, 62473271), the Beijing Natural Science Foundation (Grant No. QY24214), and the BUPT Innovation and Entrepreneurship Support Program (Grant Nos. 2025-YC-A033, 2025-YC-A042). This work is also supported by the Engineering Research Center of Information Networks, Ministry of Education, China. We would also like to thank the anonymous reviewers and area chairs for constructive discussions and feedback.
% \bibliography{anthology, custom, llms}

\appendix
\clearpage
\twocolumn
\section{Construction of FinanceReasoning}
To rigorously assess financial numerical reasoning in LLMs and LRMs, we constructed FinanceReasoning, a benchmark comprising \textbf{2,238 problems}. This dataset integrates two key sources: 

\begin{itemize} [leftmargin=*]
\itemsep0em
\item \textbf{1,420 updated existing problems}, carefully reviewed and refined from existing datasets, involving \textbf{Disambiguation}, \textbf{Elaboration} and \textbf{Correction}.
\item \textbf{908 newly generated problems}, synthesized by LLMs (\ie{ GPT-4o}) using a financial function library that we collected and open-sourced, containing 3,133 Python-formatted financial functions. Each generated problem is accompanied by an executable Python solution and precise numerical answer, all expert-verified to ensure accuracy and robustness.
\end{itemize}

To systematically evaluate models' capabilities, we introduce a tiered classification of difficulty based on computational complexity and reasoning depth:
\begin{itemize} [leftmargin=*]
\itemsep0em
\item \emph{Easy}: Direct data extraction and minimal computation, involving simple value retrieval with $\leq$2-step reasoning (\eg{ YoY growth rate calculation}).
\item \emph{Medium}:  Basic percentage operations and standard financial formula application without structural modifications, requiring $\geq$3-step computations with multiple variables (\eg{ Lorenz curve area calculation needing 35 precise steps across 5 segments with 4 boundary checks}).
\item \emph{Hard}: Multi-stage weighted computations and complex financial concept integration, demanding $\sim$8-step reasoning with 10+ variables (\eg{ 80-10-10 mortgage calculations requiring 22+ chained operations}).
\end{itemize}

This construction process, illustrated in \autoref{fig:overreview_v3}, underscores FinanceReasoning's \textbf{credibility}, \textbf{comprehensiveness}, and \textbf{challenge}, setting a new standard for evaluating financial numerical reasoning in AI models.

\section{Error Cases} \label{app:error_case}
For each of the four error types, we provide representative DeepSeek-R1 examples in \autoref{tab:errUnd}, \autoref{tab:errFor}, \autoref{tab:errExt}, and \autoref{tab:errCal}, respectively.

\section{Details of Public Dataset Updates} 
In this section, we present five common issues encountered in CodeFinQA (Test), CodeTAT-QA (Test), FinCode (Test), and FinanceMath (Val), along with the corresponding updating approaches. Besides, we analyze the distribution of difficulty among the public datasets and our dataset.

\subsection{Cases of Dataset Update} 

\label{app:revise_example}
\begin{itemize} [leftmargin=*]
\itemsep0em 
\item \textbf{Unsolvable Problem}:
Problems that either lack sufficient data or are inherently unsolvable. To address this, the case in \autoref{tab:process_err1} was updated by applying the \textbf{Disambiguation} strategy to refine the question.

\item \textbf{Ambiguous Statement}:
Questions or statements that are unclear or open to multiple interpretations. In the case of \autoref{tab:process_err2}, we updated the question by applying the \textbf{Disambiguation} strategy.

\item \textbf{Oversimplified Process}:
Inadequate or missing steps in problem-solving processes can make calculations or conclusions difficult to follow. To address this, we enhanced the Python solution in \autoref{tab:process_err3} by applying an \textbf{Elaboration} approach, ensuring a more detailed and structured problem-solving process.

\item \textbf{Incorrect Answer}:
Answers that are mathematically or conceptually incorrect, leading to misleading conclusions.  In \autoref{tab:process_err4}, the answer was refined by applying a \textbf{Correction} process.

\item \textbf{Relaxed Evaluation}:
Issues with evaluation standards, where answers that differ in precision (\eg{ 0.01, 10 and 1\%}) are considered equivalent.
We refined the answer in \autoref{tab:process_err5} by adopting a strict standard.
\end{itemize}

\subsection{Difficulty Distribution Analysis}\label{app:difficulty}
The distribution of difficulty among public datasets and FinanceReasoning is shown in \autoref{fig:difficulty_distribution}. As can be seen, the difficulty distribution of \ours surpasses previous datasets with respect to the number of medium and hard-level questions, thereby facilitating a more accurate reflection of the model's reasoning ability.

\section{Details of Dataset Expansion} 
\subsection{Example of Financial Function}
The constructed financial functions primarily consist of detailed computational code accompanied by comprehensive annotations. These annotations include explanations of function calculations, descriptions of input parameters and output variables, as well as specifications of application scope and constraints. This enhances the readability of the functions and strengthens the applicability of LLMs through well-structured documentation.
For detailed information, refer to \autoref{tab:exmFunc}.

\subsection{Example of Financial Problem}\label{app:annotation_case}
The expanded questions are based on the financial functions and concepts extracted earlier, ensuring that the questions are both relevant to the domain and challenging enough to test deep reasoning. For detailed information, refer to \autoref{Tab: example_generated_data}.

\section{Case of Knowledge Augmentation}\label{app:rag_case}
The comparison between the performance of \textbf{DeepSeek-R1} and \textbf{knowledge augmentation} are shown in \autoref{tab:r1+rag}. When using DeepSeek-R1 alone, errors may occur due to a lack of sufficient financial knowledge. However, when the knowledge augmentation is applied, the retrieval of external knowledge provides additional context and enables the model to generate the correct answer. 

\section{Case of Reasoner with Programmer}\label{app:reasoner_cot_case}
The comparison between the performance of \textbf{DeepSeek-R1 combined with Claude 3.5 Sonnet} and \textbf{DeepSeek-R1 alone} are shown in \autoref{example:r1+claude}. The task involves solving a problem that, when attempted using DeepSeek-R1 alone, results in errors. However, when DeepSeek-R1 is combined with Claude 3.5 Sonnet, the solution is correctly generated. This comparison highlights the advantage of using reasoning-enhanced model in combination with code-specialized model to improve the accuracy and reliability of the model's output.

\section{More Experimental Details}\label{app:more_details}
\begin{figure*}[!t]
    \centering
    \includegraphics[width = \linewidth]{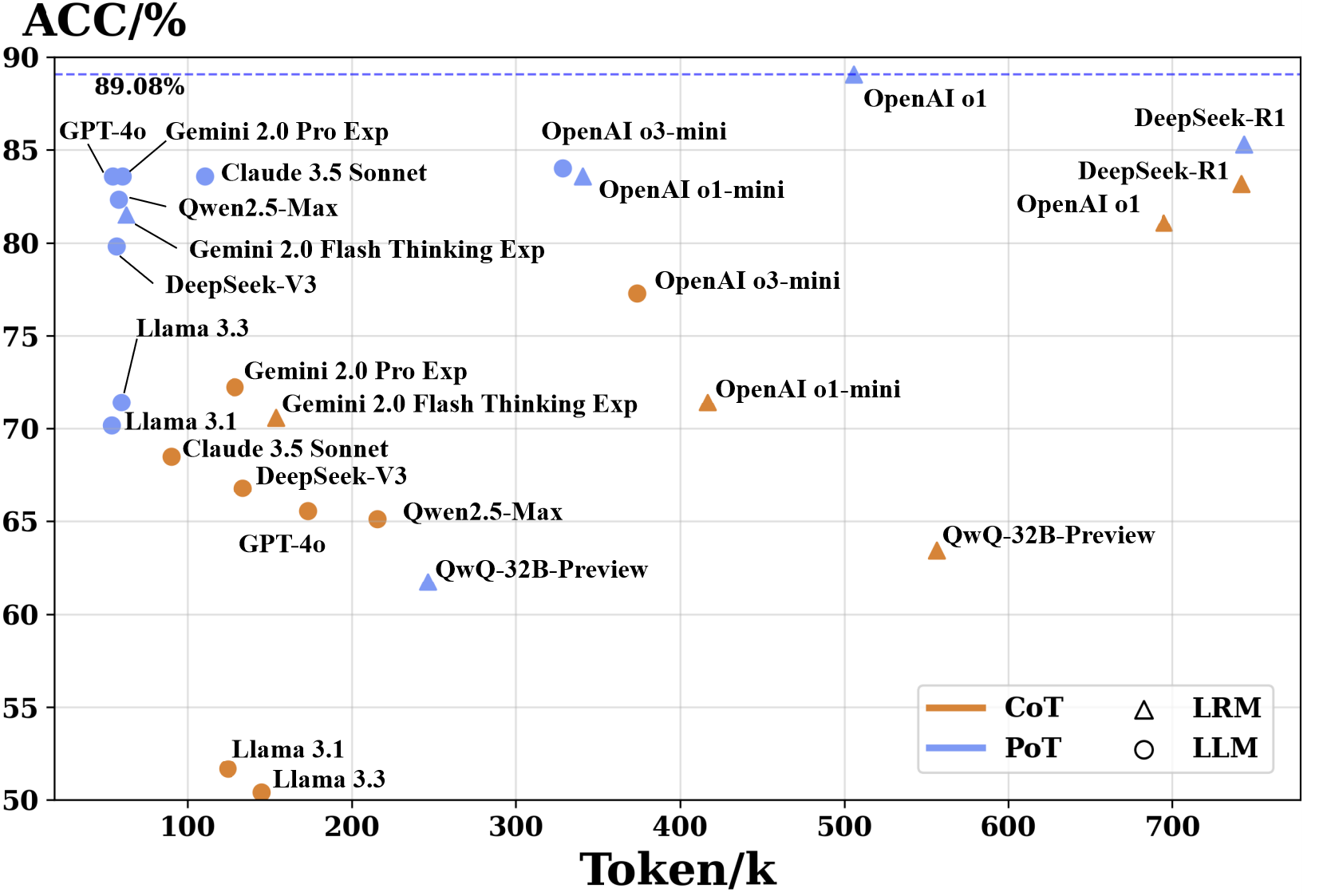}
    \caption{Relationship between token consumption and accuracy across different LLMs and LRMs using CoT and PoT prompting.}
    \label{fig:cost_effective}
\end{figure*}

The details of the evaluated models are provided in \autoref{tab:model-detail}. The token consumption of LLMs and LRMs in CoT and PoT settings is provided in \autoref{tab:tokens}. Additionally, the relationship between token consumption and accuracy across the evaluated models is illustrated in \autoref{fig:cost_effective}.

\begin{table*}[!t]
\centering
\begin{adjustbox}{max width=\linewidth}
\begin{tabular}{@{}p{\linewidth}@{}}
\toprule
\textbf{Question:}
A risk manager specializing in market risk is looking to determine the cost of a 2-year zero-coupon bond. The current 1-year interest rate is 8.0\%. There's a chance of 50\% that the 1-year interest rate will reach 10.0\% after one year and an equal probability of its falling to 6.0\% in the same period. Suppose the yearly risk premium for duration risk is 40 bps, and the bond faces a value of EUR 1,000. What should be the zero-coupon bond's valuation? Answer to three decimal places. \\
\midrule
\textbf{Question ID:} test-2164 \\
\textbf{Ground Truth:} \hlgreen{854.466} \\
\textbf{Python Solution:} \\

\texttt{def solution():} \\
\texttt{\ \ \ \ bond\_face\_value = 1000} \\
\texttt{\ \ \ \ risk\_free\_rate = 0.08} \\
\texttt{\ \ \ \ risk\_premium = 0.004} \\
\texttt{\ \ \ \ up\_interest\_rate = 0.10} \\
\texttt{\ \ \ \ down\_interest\_rate = 0.06} \\
\texttt{\ \ \ \ probability\_up = 0.5} \\
\texttt{\ \ \ \ probability\_down = 0.5} \\
\texttt{\ \ \ \ } \\
\texttt{\ \ \ \ up\_price\_date1 = bond\_face\_value / (1 + up\_interest\_rate + risk\_premium)} \\
\texttt{\ \ \ \ down\_price\_date1 = bond\_face\_value / (1 + down\_interest\_rate + risk\_premium)} \\
\texttt{\ \ \ \ } \\
\texttt{\ \ \ \ expected\_price\_date1 = probability\_up * up\_price\_date1 + probability\_down * down\_price\_date1} \\
\texttt{\ \ \ \ zero\_coupon\_bond\_price = expected\_price\_date1 / (1 + risk\_free\_rate)} \\
\texttt{\ \ \ \ } \\
\texttt{\ \ \ \ return zero\_coupon\_bond\_price}\\

\midrule
\textbf{DeepSeek-R1:}\\
To determine the valuation of the 2-year zero-coupon bond, we account for the uncertain future interest rates and the yearly risk premium of 40 basis points (bps). The steps are as follows:\\
\dots \\
3. Discounting to Present Value: \\
The expected value at \( t = 1 \) is discounted back to \( t = 0 \) using \textcolor{red}{the first-year rate adjusted for the risk premium (\( 8.0\% + 0.4\% = 8.4\% \))}:
\[
\text{PV}_0 = \frac{922.8234}{1 + 0.084} \approx 851.2993
\]

Therefore, the answer is \hlred{851.299}.\\
\bottomrule
\end{tabular}
\end{adjustbox}
%\caption{Misunderstanding of Problem Error Example by DeepSeek-R1. The wrong answer is highlighted in \hlred{red}, the ground truth is highlighted in \hlgreen{green}, while the incorrectly generated code is shown in \textcolor{red}{red}. The model misunderstood the current 1-year interest rate without risk premium.}
\caption{Misunderstanding of Problem Example by DeepSeek-R1. The wrong answer is highlighted in \protect\hlred{red}, the ground truth is highlighted in \protect\hlgreen{green}, while the incorrectly generated content is shown in \protect\textcolor{red}{red}. At \( t = 0 \), there is no need for risk premium adjustment other than that related to interest rate uncertainty, while the model incorrectly adjusted the discount rate at \( t = 0 \) (by incorporating a risk premium).}
\label{tab:errUnd}
\end{table*}
\begin{table*}[!t]
\centering
\begin{adjustbox}{max width=\linewidth}
\begin{tabular}{@{}p{\linewidth}@{}}
\toprule
\textbf{Question: }
Smith demonstrates to Harmonica the total return of a recent transaction performed by the Zenith Fund. Smith explains that the Zenith Fund entered a fully collateralized long position in nearby soybean futures contracts at the quoted futures price of 865.0 (US cents/bushel). Three months afterward, the whole futures position was rolled when the short-term futures price was 877.0 and the long-term futures price was 883.0. What is the Zenith Fund's three-month total return on the soybean futures transaction, factoring in that the collateral earned an annualized rate of 0.60\% during the three-month period between when the initial long position was established and the rolling of the contract (expressed as a percentage)? Answer to three decimal places. \\
\midrule
\textbf{Question ID:} test-2162 \\
\textbf{Ground Truth:} \hlgreen{0.853} \\
\textbf{Python Solution:} \\
\texttt{def solution():} \\
\texttt{\ \ \ \ \dots} \\
\texttt{\ \ \ \ \# Calculating each return} \\
\texttt{\ \ \ \ price\_return = (current\_price\_short\_term - previous\_price) / previous\_price} \\
\texttt{\ \ \ \ roll\_return = (current\_price\_short\_term - current\_price\_long\_term) / }\\

\texttt{current\_price\_short\_term} \\
\texttt{\ \ \ \ collateral\_return = (months / 12) * (collateral\_rate / 100)} \\
\texttt{} \\
\texttt{\ \ \ \ \# Calculating total return} \\
\texttt{\ \ \ \ total\_return = price\_return + roll\_return + collateral\_return} \\
\texttt{} \\
\texttt{\ \ \ \ return total\_return * 100}\\

\midrule
\textbf{DeepSeek-R1:}\\
The Zenith Fund's total return consists of the gain from the futures position and the interest earned on the collateral.\\

1. Futures Return: \\
The futures price increased from 865.0 to 877.0 over three months.
\[\text{Profit} =  877.0 - 865.0 = 12.0 \ \text{cents/bushel}\]
\[\text{Return} =  \frac{12.0}{865.0} = 0.013872\ (\text{or 1.3872\%}) \]

2. Collateral Return: \\
The annualized interest rate is 0.60\%, so the three-month rate is:
\[ 0.60\% \times \frac{3}{12} = 0.15\% \]
3. Sum of futures return and collateral return:
\textcolor{red}{
\[
1.3872\% + 0.15\% = 1.5372\%
\]}
Rounded to three decimal places, the total return is \textbf{1.537\%}.\\
Therefore, the answer is \hlred{1.537}.\\ 
\bottomrule
\end{tabular}
\end{adjustbox}
%\caption{Formula Application Error Example by DeepSeek-R1. The wrong answer is highlighted in \hlred{red}, the ground truth is highlighted in \hlgreen{green}, while the incorrectly generated code is shown in \textcolor{red}{red}. The model lacks knowledge of the FICO score, leading to an incorrect formula.}
\caption{Formula Application Error Example by DeepSeek-R1. The wrong answer is highlighted in \protect\hlred{red}, the ground truth is highlighted in \protect\hlgreen{green}, while the incorrectly generated content is shown in \protect\textcolor{red}{red}. The model failed to account for the roll yield component in the total return calculation, resulting in an overestimated final answer.}
\label{tab:errFor}
\end{table*}
\begin{table*}[!t]
\centering
\begin{adjustbox}{max width=\linewidth}
\begin{tabular}{@{}p{\linewidth}@{}}
\toprule
\textbf{Question:}
What is the taxable income from Jane's ESPP transaction? Answer to the nearest dollar. \\
\midrule
\textbf{Question ID:} test-2017\\
\textbf{Ground Truth:} \hlgreen{4200} \\
\textbf{Python Solution:} \\
\texttt{def solution():} \\

\texttt{\ \ \ \ \# Define the given variables} \\
\texttt{\ \ \ \ purchase\_price = 40.0  \# Purchase price per share in dollars} \\
\texttt{\ \ \ \ sale\_price = 55.0  \# Sale price per share in dollars} \\
\texttt{\ \ \ \ shares = 200  \# Number of shares} \\
\texttt{\ \ \ \ discount = 0.15  \# ESPP discount} \\
\texttt{\ \ \ \ holding\_period = 1.5  \# Holding period in years (less than 2)} \\
\texttt{\ \ \ \ } \\
\texttt{\ \ \ \ \# Calculate the discounted purchase price per share} \\
\texttt{\ \ \ \ discounted\_purchase\_price = purchase\_price * (1 - discount)} \\
\texttt{\ \ \ \ } \\
\texttt{\ \ \ \ \# Calculate the total gain from selling the shares} \\
\texttt{\ \ \ \ total\_gain = (sale\_price - discounted\_purchase\_price) * shares} \\
\texttt{\ \ \ \ } \\
\texttt{\ \ \ \ \# Since the holding period is less than 2 years, treat the entire gain as ordinary income} \\
\texttt{\ \ \ \ taxable\_income = total\_gain} \\
\texttt{\ \ \ \ } \\
\texttt{\ \ \ \ \# Return the answer rounded to the nearest dollar} \\
\texttt{\ \ \ \ return round(taxable\_income)} \\

\textbf{Context:}  \\
Jane participates in her company's Employee Stock Purchase Plan (ESPP), which offers a 15\% discount on stock purchases. She bought 200 shares at a purchase price of \$40 per share. After holding the shares for 18 months, she decides to sell them at \$55 per share. The IRS mandates that any share held less than two years must have its gain treated entirely as ordinary income. Jane needs to calculate the taxable income from this transaction to determine her tax liability. \\
\midrule
\textbf{DeepSeek-R1:}\\
Jane's taxable income from the ESPP transaction involves two components: the discount received at purchase and the gain from selling the shares, both treated as ordinary income due to the holding period being less than two years.

1. Discount Calculation (Compensation Income):\\
- The purchase price per share (\$40) reflects a 15\% discount. The fair market value (FMV) at purchase is calculated as: 
\[
\text{FMV at purchase} = \frac{\$40}{1 - 0.15} = \frac{\$40}{0.85} \approx \$47.06 \text{ per share}.
\]
- Discount per share:  
\[
\textcolor{red}{\$47.06} - \$40 = \$7.06.
\]
\dots\\

Therefore, the answer is \hlred{3000}.\\ 
\bottomrule
\end{tabular}
\end{adjustbox}
%\caption{Numerical Extraction Error Example by DeepSeek-R1. The wrong answer is highlighted in \hlred{red}, the ground truth is highlighted in \hlgreen{green}, while the incorrectly generated code is shown in \textcolor{red}{red}. The model treated the purchase price (pre-discount price) as the post-discount price.}
\caption[Numerical Extraction Error Example by DeepSeek-R1.]{Numerical Extraction Error Example by DeepSeek-R1. The wrong answer is highlighted in \hlred{red}, the ground truth is highlighted in \hlgreen{green}, while the incorrectly generated content is shown in \textcolor{red}{red}. The model treated the purchase price (pre-discount price) as the post-discount price.}
\label{tab:errExt}
\end{table*}
\begin{table*}[!t]
\centering
\begin{adjustbox}{max width=\linewidth}
\begin{tabular}{@{}p{\linewidth}@{}}
\toprule
\textbf{Question:} 
What is the fair price of the European put option? Answer in USD, rounded to two decimal places. \\
\midrule
\textbf{Question ID:} test-2208 \\
\textbf{Ground Truth:} \hlgreen{5.46} \\
\textbf{Python Solution:} \\
\texttt{def solution():} \\
\texttt{\ \ \ \ \dots} \\
\texttt{\ \ \ \ \# Define variables} \\
\texttt{\ \ \ \ S = 150  \# Current stock price in USD} \\
\texttt{\ \ \ \ K = 140  \# Strike price in USD} \\
\texttt{\ \ \ \ r = 0.02  \# Risk-free interest rate (as a decimal)} \\
\texttt{\ \ \ \ T = 0.5  \# Time to maturity in years} \\
\texttt{\ \ \ \ volatility = 0.25  \# Volatility as a decimal} \\
\texttt{\ \ \ \ \# Calculate d1 and d2 using the Black-Scholes formula} \\
\texttt{\ \ \ \ d1 = (log(S / K) + (r + 0.5 * volatility ** 2) * T) / (volatility * sqrt(T))} \\
\texttt{\ \ \ \ d2 = d1 - volatility * sqrt(T)} \\
\texttt{\ \ \ \ \# Calculate the put option price} \\
\texttt{\ \ \ \ put\_price = K * exp(-r * T) * norm.cdf(-d2) - S * norm.cdf(-d1)} \\
\texttt{\ \ \ \ \# Return the put price rounded to two decimal places} \\
\texttt{\ \ \ \ return round(put\_price, 2)} \\

\textbf{Context:}  \\
A professional investor is considering purchasing a European put option for a technology stock currently trading at 150 USD. The put option has the following parameters: \\ 
Current stock price, \( S = 150 \) USD \\
Strike price, \( K = 140 \) USD \\
Time to maturity, \( T = 0.5 \) years \\
Risk-free interest rate, \( r = 2\% \) per annum \\
Volatility, \( \sigma = 25\% \) per annum \\

The investor wants to determine the fair price of the put option to make an informed investment decision.\\

\midrule
\textbf{DeepSeek-R1:}\\
\dots\\
2. Calculate \( d_1 \) and \( d_2 \):\\
We use the following formulas for \( d_1 \) and \( d_2 \):

\[
d_1 = \frac{\ln(S/K) + (r + \frac{\sigma^2}{2}) T}{\sigma \sqrt{T}} = \frac{\ln(150/140) + (0.02 + 0.25^2/2) \cdot 0.5}{0.25 \sqrt{0.5}}
\]

\textcolor{red}{\( \ln(150/140) \approx 0.06908 \)}\\
\( (0.02 + 0.03125) \cdot 0.5 = 0.025625 \) \\
Numerator: \( 0.06908 + 0.025625 = 0.094705 \) \\
Denominator: \textcolor{red}{\( 0.25 \cdot \sqrt{0.5} \approx 0.1767767 \)}
\textcolor{red}{ \( d_1 \approx \frac{0.094705}{0.1767767} \approx 0.5357 \)} \\
Next, calculate \( d_2 \):
\[
\textcolor{red}{d_2 = d_1 - \sigma \sqrt{T} = 0.5357 - 0.1767767 \approx 0.3589} 
\]
\dots\\
After refining calculations with precise approximations and standard normal CDF values, the fair price of the European put option rounds to \hlred{5.44} USD.\\
\bottomrule
\end{tabular}
\end{adjustbox}
\caption{Numerical Calculation Error Example by DeepSeek-R1. The wrong answer is highlighted in \protect\hlred{red}, the ground truth is highlighted in \protect\hlgreen{green}, while the incorrectly generated content is shown in \protect\textcolor{red}{red}. The model applied excessive rounding during the calculation steps.}
\label{tab:errCal}
\end{table*}

\begin{table*}[!t]
\centering
\begin{adjustbox}{max width=\linewidth}
\begin{tabular}{@{}p{\linewidth}@{}}
\toprule
\textbf{Question: }As of december 31 , \hlred{2006} , what was the total total cash obligations aggregate carrying value of long-term debt due in \hlred{2006} Answer to three decimal places.\\
\midrule
\textbf{The Revised Question: }As of December 31, \hlgreen{2011}, what was the total cash obligations aggregate carrying value of long-term debt due in \hlgreen{2011}? Answer as a percentage to three decimal places.\\
\midrule
\textbf{Question ID:} test-114 \\
\textbf{Ground Truth:} 9.562 \\
\textbf{Python Solution:}\\
\texttt{def solution():} \\
\texttt{\ \ \ \ notes\_2011 = 338501} \\
\texttt{\ \ \ \ total\_obligations = 3540009} \\
\texttt{\ \ \ \ percent\_2011 = notes\_2011 / total\_obligations} \\
\texttt{\ \ \ \ answer = percent\_2011 * 100} \\
\texttt{\ \ \ \ return round(answer, 3)} \\
\textbf{Context:}  \\
\dots\\
\subsection*{Maturities 2014}

As of December 31, 2006, the aggregate carrying value of long-term debt, including capital leases, for the next five years and thereafter are estimated to be (in thousands):
\begin{center}
\begin{tabular}{|c|c|}
\hline
\textbf{Year Ending December 31} & \textbf{Amount (\$)} \\
\hline
2007 & 253,907 \\
2008 & 1,278 \\
2009 & 654 \\
2010 & 1,833,416 \\
2011 & 338,501 \\
Thereafter & 1,112,253 \\
Total Cash Obligations & 3,540,009 \\
Accreted Value of the Discount and Premium of 3.00\% Notes and 7.125\% Notes & 3,007 \\
Balance as of December 31, 2006 & 3,543,016 \\
\hline
\end{tabular}
\end{center}
The holders of the company's 5.0\% (5.0\%) notes have the right to require the company to repurchase their notes on specified dates prior to the maturity date in 2010, but the company may pay the purchase price by issuing shares of Class A common stock, subject to certain conditions. Obligations with respect to the right of the holders to put the 5.0\% (5.0\%) notes have been included in the table above as if such notes mature the date on which the put rights become exercisable in 2007. In February 2007, the company conducted a cash tender offer for its outstanding 5.0\% (5.0\%) notes to enable note holders to exercise their right to require the company to purchase their notes. (See Note 19.)\\
\dots\\
\bottomrule
\end{tabular}
\end{adjustbox}
\caption{Example of Unsolvable Problem in CodeFinQA. The parts highlighted in \protect\hlred{red} are unsolvable , and the parts highlighted in \protect\hlgreen{green} are the revised ones. Since the question is flawed due to missing 2006 data in the table, it cannot be solved, making the entire output meaningless.}
\label{tab:process_err1}
\end{table*}
\begin{table*}[!t]
\centering
\begin{adjustbox}{max width=\linewidth}
\begin{tabular}{@{}p{\linewidth}@{}}
\toprule
\textbf{Question:} \hlred{What is the change} in total cost of revenue between 2019 and 2018? Answer to two decimal places.\\
\midrule
\textbf{The Revised Question:} \hlgreen{What is the percentage change} in total cost of revenue between 2019 and 2018? Answer as a percentage to two decimal places. \\
\midrule
\textbf{Question ID:} test-142 \\
\textbf{Ground Truth:} -5.81 \\
\textbf{Python Solution:} \\
\texttt{def solution():} \\
\texttt{\ \ \ \ \dots} \\
\texttt{\ \ \ \ total\_cost\_of\_revenue\_2019 = df["Cost of revenue: -- Total cost of revenue"]["2019"]} \\
\texttt{\ \ \ \ total\_cost\_of\_revenue\_2018 = df["Cost of revenue: -- Total cost of revenue"]["2018"]} \\
\texttt{\ \ \ \ answer = ( total\_cost\_of\_revenue\_2019 - total\_cost\_of\_revenue\_2018 ) / total\_cost\_of\_revenue\_2018 * 100.0} \\
\texttt{\ \ \ \ return round(answer, 2)} \\
\textbf{Context:}
\begin{center}
\begin{tabular}{|c|c|c|c|c|}
\hline
\textbf{Category} & \textbf{2019} & \textbf{2018} & \textbf{Amount} & \textbf{Percent} \\
\hline
Cost of revenue: -- Products & 29816 & 34066 & -4250 & -12\% \\
Cost of revenue: -- Services & 19065 & 17830 & 1235 & 7\% \\
Cost of revenue: -- Total cost of revenue & 48881 & 51896 & -3015 & -6\% \\
\hline
\end{tabular} 
\end{center} \\

\bottomrule
\end{tabular}
\end{adjustbox}
\caption{Example of Ambiguous Statement in CodeTAT-QA. The parts highlighted in \protect\hlred{red} are ambiguous, and the parts highlighted in \protect\hlgreen{green} are the revised ones. The ambiguous phrasing of the question leads to multiple possible answers.}
\label{tab:process_err2}
\end{table*}
\begin{table*}[!t]
\centering
\begin{adjustbox}{max width=\linewidth}
\begin{tabular}{@{}p{\linewidth}@{}}
\toprule
\textbf{Question:} Assuming that the variances of the underlying populations are equal, independent samples taken from normally distributed groups display the following features: The combined estimation of the common variance is 2,678.05. What is the suitable t-test statistic to verify the assumption that the average of the two populations are the same? \\
\midrule
\textbf{Question ID:} validation-24 \\
\textbf{Ground Truth:} 0.938 \\
\textbf{Topic:} Quantitative Analysis \& Valuation \\
\textbf{\hlred{Python Solution:}} \\
\texttt{def solution():} \\
\texttt{\ \ \ \ return (200 - 185) / (2678.05 / 25 + 2678.05 / 18)**0.5} \\
\textbf{\hlgreen{The Revised Python Solution:}} \\
\texttt{def solution():} \\
\texttt{\ \ \ \ \# Given values} \\
\texttt{\ \ \ \ sample\_size\_A = 25} \\
\texttt{\ \ \ \ sample\_size\_B = 18} \\
\texttt{\ \ \ \ sample\_mean\_A = 200} \\
\texttt{\ \ \ \ sample\_mean\_B = 185} \\
\texttt{\ \ \ \ combined\_variance = 2678.05} \\
\texttt{\ \ \ \ } \\
\texttt{\ \ \ \ \# Standard error of the difference in means} \\
\texttt{\ \ \ \ standard\_error = (combined\_variance / sample\_size\_A + combined\_variance / sample\_size\_B)**0.5} \\
\texttt{\ \ \ \ } \\
\texttt{\ \ \ \ \# t-test statistic calculation} \\
\texttt{\ \ \ \ t\_statistic = (sample\_mean\_A - sample\_mean\_B) / standard\_error} \\
\texttt{\ \ \ \ } \\
\texttt{\ \ \ \ return round(t\_statistic, 3)} \\
\textbf{Context:}
\begin{center}
\begin{tabular}{|c|c|c|c|}
\hline
\textbf{Sample Name} & \textbf{Sample Size} & \textbf{Sample Mean} & \textbf{Sample Standard Deviation} \\
\hline
A & 25 & 200 & 45 \\
B & 18 & 185 & 60 \\
\hline
\end{tabular} 
\end{center} \\
\bottomrule
\end{tabular}
\end{adjustbox}
\caption{Example of Oversimplified Process in FinanceMath. The parts highlighted in \protect\hlred{red} are oversimplified, and the parts highlighted in \protect\hlgreen{green} are the revised ones. While the answer is correct, the solution process lacks clarity, and the code omits some steps.}
\label{tab:process_err3}
\end{table*}
\begin{table*}[!t]
\centering
\begin{adjustbox}{max width=\linewidth}
\begin{tabular}{@{}p{\linewidth}@{}}
\toprule
\textbf{Question:} An investor purchases a five-year, 6\% annual-coupon payment bond at 89.4535 and sells it in four years. Following the purchase of the bond and prior to the receipt of the first coupon, interest rates go down to 8.5\%. What is the realized yield on the bond investment? Answer as a percentage with two decimal places. \\
\midrule
\textbf{\hlred{Before Update}}\\
\textbf{Question ID:} test-8 \\
\textbf{\hlred{Answer:}} 3.7699999809 \\
\textbf{\hlred{Python Solution:}} \\
\texttt{def solution():} \\
\texttt{\ \ \ \ coupon\_rate = 0.06} \\
\texttt{\ \ \ \ initial\_price = 89.4535} \\
\texttt{\ \ \ \ years\_til\_maturity = 5} \\
\texttt{\ \ \ \ interest\_rate = 0.085} \\
\texttt{\ \ \ \ coupon = coupon\_rate * initial\_price} \\
\texttt{\ \ \ \ discount = initial\_price - coupon} \\
\texttt{\ \ \ \ final\_price = discount * (1 + interest\_rate) ** (years\_til\_maturity - 4)} \\
\texttt{\ \ \ \ realized\_yield = (final\_price / initial\_price) - 1} \\
\texttt{\ \ \ \ realized\_yield\_per = 100.0 * realized\_yield} \\
\texttt{\ \ \ \ return round(realized\_yield\_per, 2)} \\
\midrule
\textbf{\hlgreen{After Update}}\\
\textbf{Question ID:} test-8 \\
\textbf{\hlgreen{Answer:}} 8.71 \\
\textbf{\hlgreen{Python Solution:}} \\
\texttt{def solution():} \\
\texttt{\ \ \ \ face\_value = 100} \\
\texttt{\ \ \ \ coupon\_rate = 0.06} \\
\texttt{\ \ \ \ initial\_price = 89.4535} \\
\texttt{\ \ \ \ years\_til\_maturity = 5} \\
\texttt{\ \ \ \ interest\_rate = 0.085} \\
\texttt{\ \ \ \ holding\_period = 4} \\
\texttt{\ \ \ \ coupon = coupon\_rate * face\_value} \\
\texttt{\ \ \ \ sell\_price = (face\_value + coupon) / ((1 + interest\_rate) ** (years\_til\_maturity - holding\_period))} \\
\texttt{\ \ \ \ coupon\_received = coupon + coupon * (1 + interest\_rate) + coupon * (1 + interest\_rate) ** 2 + coupon * (1 + interest\_rate) ** 3} \\
\texttt{\ \ \ \ realized\_yield = ((sell\_price + coupon\_received) / initial\_price) ** (1 / holding\_period) - 1} \\
\texttt{\ \ \ \ realized\_yield\_per = 100.0 * realized\_yield} \\
\texttt{\ \ \ \ return round(realized\_yield\_per, 2)} \\

\bottomrule
\end{tabular}
\end{adjustbox}
\caption{Example of Incorrect Answer in FinCode. The case had an incorrect answer and incorrect python solution, which we have corrected. The parts highlighted in \protect\hlred{red} are incorrect, and the parts highlighted in \protect\hlgreen{green} are the revised ones.}
\label{tab:process_err4}
\end{table*}
\begin{table*}[!t]
\centering
\begin{adjustbox}{max width=\linewidth}
\begin{tabular}{@{}p{\linewidth}@{}}
\toprule
\textbf{Question:} What is the holding period return for the three-year timeframe, given the following annual rates of return for a mutual fund as reported by a researcher (expressed as a percentage)? \\
\midrule
\textbf{The Revised Question:} What is the holding period return for the three-year timeframe, given the following annual rates of return for a mutual fund as reported by a researcher (expressed as a percentage)? \hlgreen{Answer to three decimal places.}\\
\midrule
\textbf{Question ID:} validation-68 \\
\textbf{Ground Truth:} 0.548 \\
\textbf{Python Solution:} \\
\texttt{def solution():} \\
\texttt{\ \ \ \ return round(((1+0.14)*(1-0.10)*(1-0.02)-1)*100, 3)} \\
\textbf{Context:}
\begin{center}
\begin{tabular}{|c|c|}
\hline
\textbf{Year} & \textbf{Return (\%)} \\
\hline
2008 & 14 \\
2009 & -10 \\
2010 & -2 \\
\hline
\end{tabular} 
\end{center} \\
\bottomrule
\end{tabular}
\end{adjustbox}

\caption{Example of Relaxed Evaluation in FinanceMath. The parts highlighted in \protect\hlgreen{green} are the revised ones. The question is answered with an evaluation that lacks rigor, and multiple correct answers are possible.}
\label{tab:process_err5}
\end{table*}

\begin{table*}[!t]
\centering
\begin{adjustbox}{max width=\linewidth}
\begin{tabular}{@{}p{\linewidth}@{}}
\toprule
\textbf{Financial Function:} \\
\begin{lstlisting}[language=Python]
def calculate_hedge_fund_net_return(initial_investment: float, gross_return_rate: float, management_fee_rate: float, incentive_fee_rate: float, hurdle_rate: float) -> float:
    """
    Calculate the net return for an investor in a hedge fund with fees and hurdle rates. 
    
    This function computes the net return in dollar terms for an investor after accounting for management fees and incentive fees, based on a hurdle rate.

    Args:
        initial_investment (float): The initial investment amount in millions.
        gross_return_rate (float): The gross return rate achieved by the hedge fund.
        management_fee_rate (float): The management fee rate as a decimal.
        incentive_fee_rate (float): The incentive fee rate as a decimal.
        hurdle_rate (float): The hurdle rate as a decimal percentage of initial investment.
    Returns:
        net_return (float): The net return for the investor in millions to two decimal places.
    Notes:
        - Applicability: Suitable for hedge funds using a similar fee structure.
        - Constraints: Assumes management fee is based on year-end assets and incentive fee is calculated net of management fee, based on returns over the hurdle rate.
        - Considerations: Ensure all rates are provided as decimals (e.g., 2% as 0.02).
    """
    year_end_assets = initial_investment * (1 + gross_return_rate)
    management_fee = year_end_assets * management_fee_rate
    net_year_end_assets = year_end_assets - management_fee
    hurdle_amount = initial_investment * (1 + hurdle_rate)
    excess_return = max(0, net_year_end_assets - hurdle_amount)
    incentive_fee = excess_return * incentive_fee_rate
    total_fees = management_fee + incentive_fee
    net_return = year_end_assets - total_fees - initial_investment
    
    return round(net_return, 2)
\end{lstlisting} \\

\toprule
\end{tabular}
\end{adjustbox}
\caption{Example of financial function from function library.}
\label{tab:exmFunc}
\end{table*}

\begin{table*}[!t]
\centering
\begin{adjustbox}{max width=\linewidth}
\begin{tabular}{@{}p{\linewidth}@{}}
\toprule
\textbf{Question:} What is the company's Weighted Average Cost of Capital (WACC)? Answer as a percentage to two decimal places. \\
\textbf{Question ID:} test-2002 \\
\textbf{Ground Truth:} \hlgreen{6.9} \\
\textbf{Context:} \\
A manufacturing company is evaluating its financing strategy and needs to calculate its Weighted Average Cost of Capital (WACC) to optimally structure its capital resources. The company's current market value of equity is 150 million, and the market value of its debt is 100 million. The cost of equity is estimated at 9\%, while the cost of debt stands at 5\%. Considering the corporate tax rate is 25\%, the company wants to determine its WACC to make informed investment decisions.\\

\midrule
\textbf{Function used as an instruction:}\\

\begin{lstlisting}[language=Python]
def calculate_wacc(market_value_equity: float, market_value_debt: float, cost_of_equity: float, cost_of_debt: float, tax_rate: float) -> float:
    """
    Calculate the Weighted Average Cost of Capital (WACC). 
    
    This function calculates the WACC, which represents a firm's average cost of financing from all sources, weighted by their respective usage in the overall capital structure. It gives an overall measure of the firm's cost of capital.

    Args:
        market_value_equity (float): Market value of the firm's equity.
        market_value_debt (float): Market value of the firm's debt.
        cost_of_equity (float): Cost of equity (Re) represented as a decimal (e.g., 0.08 for 8%).
        cost_of_debt (float): Cost of debt (Rd) represented as a decimal (e.g., 0.04 for 4%).
        tax_rate (float): Corporate tax rate (Tc) represented as a decimal (e.g., 0.30 for 30%).

    ...
    """
    total_value = market_value_equity + market_value_debt
    equity_weight = market_value_equity / total_value
    debt_weight = market_value_debt / total_value
    wacc = (equity_weight * cost_of_equity) + (debt_weight * cost_of_debt * (1 - tax_rate))
    
    return wacc
\end{lstlisting} \\

\bottomrule
\end{tabular}
\end{adjustbox}
\caption{A generated problem in FinanceReasoning, combined with the structured financial function as an instruction. The ground truth is highlighted in \protect\hlgreen{green}.}
\label{Tab: example_generated_data}
\end{table*}

\begin{table*}[!t]
\centering
\begin{adjustbox}{max width=\linewidth}
\begin{tabular}{@{}p{\linewidth}@{}}
\toprule
\textbf{Question:} According to the given table, can you compute the value added from the choice of security (expressed as a percentage)? Answer to a single decimal place. \\
\textbf{Question ID:} test-2193 \\
\textbf{Ground Truth:} \hlgreen{5.6} \\
\textbf{Context:} 
\begin{center}
\begin{tabular}{|c|c|c|c|c|}
\hline
\textbf{} & \textbf{Portfolio return} & \textbf{Benchmark return} & \textbf{Portfolio weight} & \textbf{Benchmark weight} \\
\hline
\textbf{Domestic} & 25\% & 15\% & 50 & 35 \\
\hline
\textbf{Bond} & 9\% & 5\% & 30 & 35 \\
\hline
\textbf{International} & 15\% & 18\% & 20 & 30 \\
\hline
\end{tabular}
\end{center}\\

\midrule
\textbf{DeepSeek-R1:}\\
To compute the value added from the choice of security (selection effect)...\\
\textbf{Domestic:}
\[(25\% - 15\%) \times \textcolor{red}{35\%} = 3.5\%
\]

\textbf{Bond:}
\[(9\% - 5\%) \times \textcolor{red}{35\%}  = 1.4\%
\]

\textbf{International:}
\[(15\% - 18\%) \times \textcolor{red}{30\%}  = -0.9\%
\]

\textbf{Total Selection Effect:}
\[
\textcolor{red}{3.5\% + 1.4\% - 0.9\% = 4.0\%}
\]

Therefore, the answer is \hlred{4.0}.\\
\midrule
\textbf{DeepSeek-R1 + Knowledge Augmentation:}\\
\textbf{Instruct from GPT-4o:}\\What function calculates the security selection effect in portfolio performance attribution...\\
\textbf{The function retrieved and considered useful by GPT-4o:}\\
\begin{lstlisting}[language=Python]
def calculate_security_selection_effect(portfolio_returns: list, benchmark_returns: list, portfolio_weights: list) -> float:
    ...
    total_security_selection_effect = sum((pr - br) * pw for pr, br, pw in zip(portfolio_returns, benchmark_returns, portfolio_weights)
    
    return round(total_security_selection_effect * 100, 1)
\end{lstlisting}

\textbf{Domestic:}
\[
(25\% - 15\%) \times \textcolor{green!50!black}{50\%} = 5.0\%
\]
\textbf{Bond:}
\[
(9\% - 5\%) \times \textcolor{green!50!black}{30\%} = 1.2\%
\]
\textbf{International:}
\[
(15\% - 18\%) \times \textcolor{green!50!black}{20\%} = -0.6\%
\]
The final selection effect is:  
\[
\textcolor{green!50!black}{5.0\% + 1.2\% - 0.6\% = 5.6\%}
\] \\

\midrule
\textbf{Output:} \textcolor{green!50!black}{5.6}\\
\bottomrule
\end{tabular}
\end{adjustbox}
\caption{ A comparison of \textbf{DeepSeek-R1} and \textbf{DeepSeek-R1 with Knowledge Augmentation} in FinanceReasoning. The ground truth is highlighted in \protect\hlgreen{green}. Segments highlighted in \protect\hlred{red} or \protect\textcolor{red}{red} indicate errors made by \textbf{DeepSeek-R1}, while those shown in  \protect\textcolor{green!50!black}{green} indicate correct answers produced by \textbf{DeepSeek-R1 with Knowledge Augmentation}.}
\label{tab:r1+rag}
\end{table*}

\begin{table*}[!t]
\centering

\begin{adjustbox}{max width=\linewidth}
\begin{tabular}{@{}p{\linewidth}@{}}
\toprule
\textbf{Question:} What is the total amount of money that John will pay towards the principal in the first 12 months of his mortgage? Answer to the nearest dollar. \\
\textbf{Question ID:} test-2224 \\
\textbf{Ground Truth:} \hlgreen{4597} \\
\textbf{Context:}
\begin{center}
\begin{tabular}{|c|c|}
\hline
\textbf{Loan Amount (\$)} & 250,000 \\
\hline
\textbf{Annual Interest Rate (\%)} & 3.75 \\
\hline
\textbf{Loan Term (Years)} & 30 \\
\hline
\end{tabular}
\end{center}\\

\midrule
\textbf{DeepSeek-R1:}\\
To determine the total amount John will pay towards the principal in the first 12 months of his mortgage, we use the following steps:\\
1. **Calculate the monthly payment (M):**  \\
Using the formula for a fixed-rate mortgage...

2. **Total principal paid after 12 months:**\\
The formula for the remaining balance \( B_{12} \) after 12 payments is:
\[
\textcolor{red}{B_{12} = 250,000 \times \frac{(1.003125)^{360} - (1.003125)^{12}}{(1.003125)^{360} - 1} \approx 245,366.50} \hspace{0.6cm} (\textcolor{green!50!black}{245,403.05})
\]
So, total principal paid is: 
\[
\textcolor{red}{250,000 - 245,366.50 = 4,633.50 \quad \text{(rounded to 4634)}}\]

3. **Validation via amortization schedule:**\\
Summing the principal portions of each monthly payment (rounded to the nearest cent) results in a total principal paid of ...\\
**Therefore, the answer is \hlred{4634}.**\\
\midrule
\textbf{DeepSeek-R1 + Claude 3.5 Sonnet:}\\
Based on the reasoning process, I'll create a Python program that calculates the total principal paid in the first 12 months using the formula method:\\

\begin{lstlisting}[language=Python]
def solution():
    # Define variables name and value
    loan_amount = 250000
    annual_rate = 0.0375
    monthly_rate = annual_rate / 12
    #...
    # Calculate remaining balance after 12 months
    remaining_balance = loan_amount * (power_n - power_k) / (power_n - 1)
    # Total principal paid is original loan amount minus remaining balance
    principal_paid = loan_amount - remaining_balance
    # ...

    return answer
\end{lstlisting} \\

\midrule
\textbf{Program Output:} \textcolor{green!50!black}{4597}\\
\bottomrule
\end{tabular}
\end{adjustbox}
\caption{A comparison of \textbf{DeepSeek-R1 alone} and \textbf{DeepSeek-R1 combined with Claude 3.5 Sonnet} in FinanceReasoning. The ground truth is highlighted in \protect\hlgreen{green}. Segments highlighted in \protect\hlred{red} or \protect\textcolor{red}{red} indicate errors by \textbf{DeepSeek-R1}, while those shown in \protect\textcolor{green!50!black}{green} indicate correct answers from \textbf{DeepSeek-R1 combined with Claude 3.5 Sonnet}.}
\label{example:r1+claude}
\end{table*}

\begin{table*}[!t]
\centering
\resizebox{1\textwidth}{!}{%
\renewcommand{\arraystretch}{1.1}
\begin{tabular}{llllp{10cm}}
\toprule
\textbf{Model} & \textbf{Organization} & \textbf{Size} & \textbf{Notes} & \textbf{Source} \\
\midrule
DeepSeek-R1 & DeepSeek & 671B & MoE &  \texttt{deepseek-ai/DeepSeek-R1} \\
DeepSeek-V3 & DeepSeek & 671B & MoE &  \texttt{deepseek-ai/DeepSeek-V3} \\
\noalign{\vskip 0.5ex}\hdashline\noalign{\vskip 0.5ex}

Claude 3.5 Sonnet & Anthropic & -- & & \texttt{claude-3.5-sonnet-1022} \\
\noalign{\vskip 0.5ex}\hdashline\noalign{\vskip 0.5ex}

Gemini 2.0 Flash Thinking Experimental & Google & -- & & \texttt{gemini-2.0-flash-thinking-exp-01-21} \\
Gemini 2.0 Pro Experimental & Google & -- & & \texttt{gemini-2.0-pro-exp-02-05} \\

\noalign{\vskip 0.5ex}\hdashline\noalign{\vskip 0.5ex}
QwQ-32B-Preview & Alibaba & 32B & & \texttt{Qwen/QwQ-32B-Preview} \\
Qwen2.5-Max & Alibaba & -- & MoE & \texttt{qwen-max-2025-01-25} \\

\noalign{\vskip 0.5ex}\hdashline\noalign{\vskip 0.5ex}
Llama 3.1 & Meta & 405B & & \texttt{meta-llama/Llama-3.1-405B-Instruct} \\
Llama 3.3 & Meta & 70B & & \texttt{meta-llama/Llama-3.3-70B-Instruct} \\

\noalign{\vskip 0.5ex}\hdashline\noalign{\vskip 0.5ex}
OpenAI o1 & OpenAI & -- & & \texttt{o1-2024-12-17} \\
OpenAI o1-mini & OpenAI & -- & & \texttt{o1-mini-2024-09-12} \\
OpenAI o3-mini & OpenAI & -- & & \texttt{o3-mini-2025-01-31} \\
GPT-4o & OpenAI & -- & & \texttt{gpt-4o-2024-11-20} \\

\bottomrule
\end{tabular}
}
\caption{Details of the model organization and model source (\ie{ model version for proprietary models, and Hugging Face model name for open-source models}) for the LLMs and LRMs evaluated in FinanceReasoning.}
\label{tab:model-detail}
\end{table*}

\begin{table*}[ht]
    \renewcommand\arraystretch{1.2} 
    \setlength{\arrayrulewidth}{0.8pt}
    \setlength{\tabcolsep}{6pt}
    \centering
    \resizebox{0.8\textwidth}{!}{%
    \begin{tabular*}{\textwidth}{@{\extracolsep{\fill}}lcc} 
    \toprule
    \multicolumn{1}{c}{\textbf{Model}} & \multicolumn{2}{c}{\textbf{Tokens (k) for Easy/Medium/Hard}} \\
    \cmidrule(lr){2-3}
    & \textbf{CoT} & \textbf{PoT} \\
    \midrule
    OpenAI o1-mini & 417 / 1,074 / 944 & 341 / 1,015 / 953 \\ 
    OpenAI o1 & 695 / 1,500 / 1,242 & 505 / - / - \\
    Gemini 2.0 Flash Thinking Experimental & 154 / 442 / 311 & 62 / 164 / 140 \\
    QwQ-32B-Preveiw & 556 / 1,514 / 1,307 & 246 / 609 / 543 \\
    DeepSeek-R1 & 742 / 1,499 / 1,274 & 743 / 1,593 / 1,257 \\
    OpenAI o3-mini & 374 / 771 / 712 & 328 / 664 / 618 \\
    
    Gemini 2.0 Pro Experimental & 128 / 295 / 206 & 60 / 143 / 112 \\
    GPT-4o & 173 / 428 / 298 & 54 / 133 / 105 \\
    Claude 3.5 Sonnet & 90 / 261 / 201 & 110 / 335 / 275 \\
    Deepseek-V3 & 133 / 322 / 237 & 56 / 144 / 114 \\
    Llama 3.3 & 145 / 350 / 263 & 60 / 156 / 130 \\
    Llama 3.1 & 125 / 318 / 235 & 53 / 139 / 109 \\
    Qwen2.5-Max & 215 / 526 / 373 & 58 / 139 / 108 \\
    \bottomrule
    \end{tabular*}
    }
    \caption{Token usage across different models.}
    \label{tab:tokens}
\end{table*}

\end{document}